\documentclass[10pt,twocolumn,letterpaper]{article}

\usepackage{cvpr}
\usepackage{times}
\usepackage{graphicx}
\usepackage{amsmath}
\usepackage{amssymb}

\usepackage[pagebackref=true,breaklinks=true,letterpaper=true,colorlinks,bookmarks=false]{hyperref}

\cvprfinalcopy 

\def\cvprPaperID{****} 

\ifcvprfinal\fi

\newcommand{\D}{\mathcal{D}}
\newcommand{\N}{\mathcal{N}}
\newcommand{\Hcal}{\mathcal{H}}
\newcommand{\Xcal}{\mathcal{X}}

\newcommand{\R}{\mathbb{R}}

\newcommand{\norm}[1]{\left\|#1\right\|}
\newcommand{\iprod}[2]{\left\langle #1, #2 \right\rangle}
\newcommand{\EE}[2]{\mathbb{E}_{#1}\left[#2\right]}
\newcommand{\ith}[1]{{#1}^\mathrm{th}}

\newcommand{\iid}{\overset{\mathrm{iid}}{\sim}}

\DeclareMathOperator{\unif}{Unif}

\newcommand{\where}{\ \mathrm{where}}

\begin{document}

\title{Deep Mean Maps}
\renewcommand{\thefootnote}{\fnsymbol{footnote}}
\author{
Junier B.\ Oliva\footnotemark[1]
\qquad  Danica J.\ Sutherland\footnotemark[1]
\qquad  Barnab\'as P\'oczos
\qquad  Jeff Schneider
\\
Carnegie Mellon University \\
\{joliva,dsutherl,bapoczos,schneide\}@cs.cmu.edu
}

\maketitle

\begin{abstract}
   The use of \emph{distributions} and \emph{high-level features from
deep architecture} has become commonplace in modern
computer vision. Both of these methodologies have
separately achieved a great deal of success in many computer vision tasks.
However, there has been little work attempting
to leverage the power of these to methodologies jointly. To
this end, this paper presents the Deep Mean Maps (DMMs) framework, a
novel family of methods to non-parametrically represent distributions
of features in convolutional neural network models.

DMMs are able to both classify images using the
distribution of top-level features, and to tune the top-level features for
performing this task. We show how to implement
DMMs using a special mean map layer composed of typical CNN
operations, making both forward and backward propagation simple.

We illustrate the efficacy of DMMs at analyzing distributional
patterns in image data in a synthetic data experiment.
We also show that we extending existing
deep architectures with DMMs improves the performance of existing
CNNs on several challenging real-world datasets.
\end{abstract}

\footnotetext[1]{These two authors contributed equally.}
\renewcommand{\thefootnote}{\arabic{footnote}}

\section{Introduction}

Modern computer vision has seen extensive use of two separate machine learning methodologies: deep architectures and learning on distributions. However, research has only just begun to consider the joint use of these two methods. In this paper we present deep mean maps (DMMs), a novel framework to jointly use and learn distributions of high-level features from deep architectures.
We show that DMMs are easy to implement into existing deep learning infrastructure, and are able to improve the results of existing architectures on various datasets.

A substantial amount of work has been devoted to constructing distribution-based features for vision tasks. These distribution-based approaches are often patch-level histogram features such as bag of words (BoW) representations  \cite{leung2001representing}, where the ``words'' are local features such as histogram of gradients (HoG) \cite{dalal2005histograms} or dense SIFT features \cite{lowe1999object}. More recent developments have extended distribution-based methods beyond histograms to nonparametric continuous domains \cite{poczos2012nonparametric,muandet2012learning}. Such methods are adept at providing robust representations that and give a holistic, aggregate view of an image helpful in complex classification tasks.

Deep architectures, which compose many layers of computational neural units, have recently achieved a tremendous amount of success in computer vision tasks by replacing hand-designed features with learned convolutional filters. In fact, deep architectures, often convolutional neural networks (CNNs), have almost become the de-facto standard in some datasets with their recent success \cite{russakovsky2014imagenet}. It is believed that the depth 
and hierarchical nature of deep CNNs lead to top-level convolutional features that exhibit semantically meaningful representations \cite{lecun2013prez}.
Typically, the last levels of these features are then concatenated and fed through a few fully-connected layers to obtain a final classification result.

It would be informative to study the distributions of these high-level features in an image for supervised tasks (e.g. scene classification), rather than treating the extracted features simply as a vector. In this paper we address the lack of the application of distributions in deep architectures through the development of a deep mean map layer, which will provide a fast and scalable featurization of top-level convolutional features that will be representative of their distributions in a nonparametric fashion.
This combination of distribution-based and deep architecture approaches has seen some initial success
with an ad-hoc method of extracting features from distributions of fixed, pre-trained high-level features \cite{wu2015visual}. However, there has yet to been work that jointly learns high-level features and uses their distributions for supervised learning tasks. To this aim we present deep mean maps, which do so in a scalable, nonparametric fashion.

Deep mean maps employ mean map embeddings \cite{gretton2006kernel,smola2007hilbert,song2008learning}, which with the use of random features \cite{rahimi2007random}, provide a finite-dimensional nonparametric representation of distributions of top-level features for discriminant learning. We will show that deep mean maps can be implemented with typical CNN machinery, making both forward and backward propagation tractable for learning effective features and discriminant models based on their distributions. 

\paragraph{Outline} The rest of the paper is structured as follows. First, we detail our DMM framework in Section~\ref{sec:method}. We then study its behavior on a synthetic problem and illustrate its efficacy on several real-world datasets in Section~\ref{sec:experiments}. Section~\ref{sec:conclusion} then discusses the relationship to other method and gives concluding remarks.

\section{Method} \label{sec:method}

In this section we explicate our framework for learning high-level descriptive features, whose distributions are used in supervised image classification problems.

We focus on deep architectures of convolutional networks, which we briefly discuss below; however, our framework extends to other architectures. We show how to treat the distributions of high-level features of convolutional networks as inputs to aid in learning tasks via mean maps \cite{smola2007hilbert}. Furthermore, we show how to learn supervised tasks using mean maps in a manner that scales to large datasets via random Fourier features \cite{rahimi2007random}. Lastly, we demonstrate that one can incorporate mean maps into a convolutional network to perform supervised tasks whilst tuning the network's high-level features for this purpose. 

\subsection{High-level features from deep architectures}
The use of deep architectures for learning high-level descriptive features has been extensively explored over the last decade \cite{hinton2006reducing}. It is believed that the deep architectures present in many modern models are instrumental to learning features for both supervised \cite{lecun2012learning,krizhevsky2012imagenet} and unsupervised \cite{lee2009convolutional,le2013building} tasks. While such high-level features may be extracted in a variety of different deep architectures, we focus on deep convolution neural networks (CNNs) \cite{lecun1995convolutional,lecun1998gradient,lecun2004learning,lecun2012learning}. 

CNNs have been used with great success in many vision tasks. Inspired by the study of visual cortices \cite{lecun1995convolutional}, CNNs use successive applications of locally-receptive convolution, subsampling, nonlinear transformation, and normalization layers. Through their localized nature, CNNs enjoy a reduction in parameters over general neural networks and may exploit the spatial structures present in image data. Furthermore, with the use of modern non-linear units such as ReLUs \cite{nair2010rectified}, efficient GPU implementations, and large datasets, modern CNNs have been able to employ more layers to achieve state of the art performance \cite{krizhevsky2012imagenet, szegedy2014going}. The high accuracy and spatial structure of deep CNN models make the outputs of top-level convolution layers excellent high-level features for image classification. Thus, we look to leverage the distribution of these features to perform supervised tasks.

\subsection{Scalable mean map embeddings}
The main tool that we will use to learn with distributions of high-level features are mean map embeddings, which we detail below. We show how mean map embeddings, along with random features \cite{rahimi2007random}, can be implemented using typical CNN layers for scalable learning of distributions of high-level features in vision tasks.

\subsubsection{Mean map embeddings}
Mean map embeddings (MMEs) of distributions serve as a way to embed distributions into a reproducing kernel Hilbert space (RKHS) \cite{gretton2006kernel,smola2007hilbert}, allowing one to apply many familiar kernel methods, such as SVMs and GPs, to datasets of distributions through the kernel trick. 
Formally, the MME of a distribution $P$ in the RKHS induced by a kernel $K(x,y) =  \iprod{\phi(X)}{\phi(y)}_{\Hcal}$ is defined as:
\begin{align}
\mu_P \equiv \EE{X \sim P}{K(X,\cdot)} = \EE{X \sim P}{\phi(X)}.
\end{align}
If $K$ is a \textit{characteristic} kernel (as is the RBF kernel), then $\mu_P$ uniquely identifies $P$ \cite{song2008learning}. Moreover, one may apply an element of the RKHS $f\in\Hcal$ to $\mu_P$ by:
\begin{align}
\iprod{f}{\mu_{P}}_\Hcal = \EE{X\sim P}{\iprod{f}{K(X,\cdot)}_\Hcal} = \EE{X\sim P}{f(X)}. \label{eq:mu-iprod}
\end{align}
Thus, given that we may compute the inner-product between the embedding of a distribution $P$ in the RKHS, $\mu_P$, and another element of the RKHS, $f$, we may set up a loss w.r.t. \eqref{eq:mu-iprod} to learn over distributions.

\subsubsection{Sample estimates}
Of course, the calculation of $\mu_P$ necessitates the population mean of $ \EE{X \sim P}{K(X,\cdot)} $, which will not be available in real-world data. Instead, one typically computes a finite sample estimate of $\mu_P$ given a sample $\Xcal = \{X_j\}_{j=1}^n \iid P$:
\begin{align}
\mu_\Xcal = \frac{1}{n} \sum_{j=1}^n K(X_j,\cdot).
\end{align}
Furthermore, one may compute $\iprod{f}{\mu_{\Xcal}}_\Hcal$ as:
\begin{align}
\iprod{f}{\mu_\Xcal}_\Hcal = \frac{1}{n} \sum_{j=1}^n f(X_j).
\end{align}
Hence, we may apply kernel methods to distributions $\{P_i\}_{i=1}^N$ when given a data-set of sample sets $\D = \{\Xcal_{i}\}_{i=1}^N$, with $\Xcal_{i} = \{ X_{ij} \}_{j=1}^{n_i}$.

\subsubsection{Random feature embeddings}
Unfortunately, due to the computation of a $N \times N$ Gram matrix of pairwise kernel evaluations, the application of kernel methods to the dataset $\D = \{\Xcal_{i}\}_{i=1}^N$ will scale as $\Omega(N^2)$~---~intractable for modern image datasets containing millions of images. To mitigate this, one may use the mean embedding of random features \cite{rahimi2007random,rahimi2009weighted} to estimate kernel evaluations in an approximate finite primal space. That is, for a shift-invariant kernel $K(x,y)=k(x-y)$ with $k:\R^d \to \R_{\geq 0}$ and $k(0)=1$ we have that
\begin{align}
&K(x,y) \approx z(x)^Tz(y) \quad \where \\
&z(x) = \tfrac{1}{\sqrt{2D}}\left( \cos( \omega_1^T x + b_1 ), \ldots, \cos( \omega_D^T x + b_D ) \right)^T,
\end{align}
$b_d \iid \unif[0,2\pi]$, and $\omega_d \iid \rho$, where $\rho$ is the spectral distribution of $k$ (i.e. its Fourier transform). Thus, we can approximate the kernel evaluation $K(x,y)$ using a finite dimensional dot-product, even for characteristic kernels induced by an infinite feature map like the RBF kernel. For the RBF kernel,
$$ K(x,y) \equiv \exp\left( -\frac{1}{2 \sigma^2} \norm{x-y}^2 \right), $$
we have that $\rho(\omega) = \N(\omega; 0, \sigma^{-2} I )$.
Furthermore, using the random feature map $z(\cdot)$ we can approximate the mean map $\mu_P$ as:
\begin{align}
\hat{\mu}_{\Xcal_i} &\equiv  \frac{1}{n} \sum_{j=1}^n z(X_j) \label{eq:hatmu}
 \approx \EE{X \sim P}{\phi(X)} =\mu_P  .
\end{align}
Moreover, we can approximate the inner product of $\mu_{P_i}$ and $f \in \Hcal$, $f(x) = \sum_{l=1}^\infty \alpha_l K(x_l,x)$ as:
\begin{align}
\iprod{f}{\mu_{P_i}}_\Hcal \approx& \frac{1}{n_i} \sum_{j=1}^{n_i} f(X_j) \\
=& \frac{1}{n_i} \sum_{j=1}^{n_i} \sum_{l=1}^\infty \alpha_l K(X_j,x_l) \\
=& \frac{1}{n_i} \sum_{j=1}^{n_i} \sum_{l=1}^\infty \alpha_l z(X_j)^T z(x_l) \\
=& \left( \frac{1}{n_i} \sum_{j=1}^{n_i} z(X_j) \right)^T \left( \sum_{l=1}^\infty \alpha_l z(x_l) \right) \\
=& \hat{\mu}_{\Xcal_i}^T \psi \label{eq:rksdot},
\end{align}
where $\psi = \sum_{l=1}^\infty \alpha_l z(x_l)$.
Hence, we see that one can estimate RKHS inner products of mean maps as finite-dimensional dot products in a primal space induced by the random features $z(\cdot)$. Thus, we may avoid the costly $\Omega(N^2)$ scaling by working in this approximate primal space. 
Below we expand on how to learn high-level features whose distributions are used in supervised tasks through mean map embeddings of random features. 

\subsection{MMEs of CNN features}

\begin{figure*}[t]
  \centering
  \includegraphics[width=.95\textwidth]{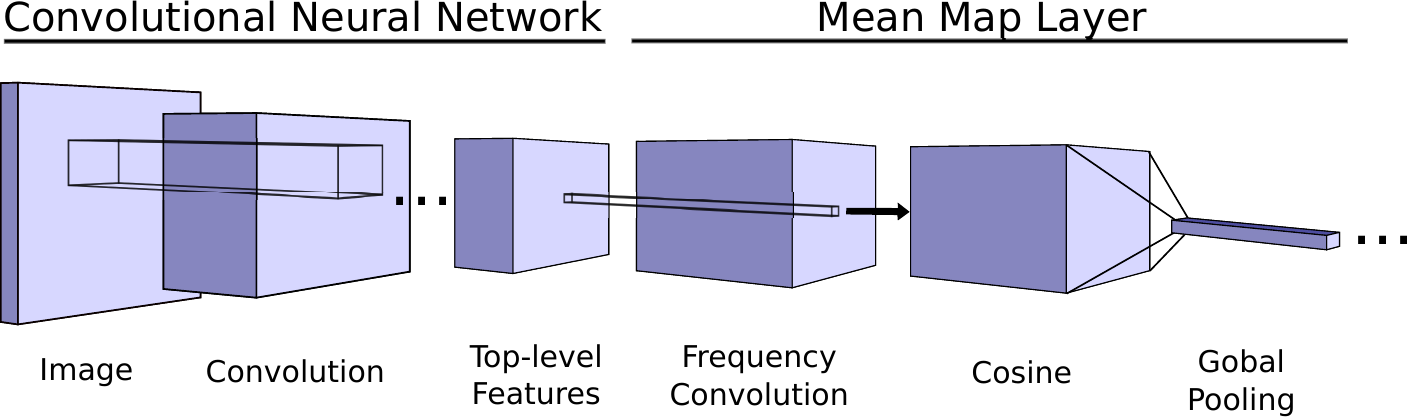}
  \caption{CNN with a mean map layer. We consider the top-level features resulting from several convolution/pooling layers. To compute the mean map embedding to the filter vectors \eqref{eq:convset}, we perform the following: 1) take $1 \times 1$ convolution with frequencies $\{ \omega_d \}_{d=1}^D$; 2) take element-wise cosines; 3) do global average pooling to compute a $D$-length vector. \label{fig:dmm}}
\end{figure*}

Given that CNNs have proven to be successful feature extractors for classification tasks, it is only natural that we look to study the distributions of said features across spatial dimensions. For many vision tasks, such as scene classification, the distribution of filter values will be highly informative, since we expect the presence of features across various locations to be relevant for such supervised tasks.
In fact, \cite{wu2015visual} achieved state-of-the-art performance whilst only considering fixed, pre-trained simple parametric approximations of the distribution of high-level features through ad-hoc summary statistics.
In a similar vein, \cite{lin2014nin} uses a final global average pooling layer on convolution features that ``sums out the spatial information," ``is more robust to spatial translations," and generalizes better. One can view this global average pooling as representing the distribution of features by its mean; our method allows richer (and, in fact, \emph{unique}) representations of these distributions.
We now illustrate how to include a layer that computes the mean map embedding of convolution filters in order to both learn informative features and use their distributions to discriminate in supervised tasks.

Suppose that the top convolution layer of a CNN (after any subsampling) produces a tensor $C_i \in \R^{m \times h \times w}$ for the $\ith{i}$ input image fed forward through the network, where here $m$ is the number of convolutional filters, and $h,w$ are spatial dimensions of super pixels (see Figure~\ref{fig:dmm}). We wish to consider the distributions of the $m \times 1$ vectors of filter values: 
\begin{align}
\Xcal_i = \left\{ C_{i,j,l}\in\R^{m} \,|\, 1 \leq j \leq h,\ 1 \leq l \leq w \right\}, \label{eq:convset}
\end{align}
where $C_{i,j,l} = \left( C_i(1,j,l), \ldots, C_i(m,j,l) \right)^T$. To do so, we will calculate the MME of $\Xcal_i$ \eqref{eq:convset} in what we term as the mean map layer. The mean map layer calculates $\hat{\mu}_{\Xcal_i}$ and propagates it forward in the network, allowing one to classify images with distributions of high-level features that are tuned for this distributional representation.

We now show that the mean map layer can be represented in terms of standard CNN mechanisms. Specifically, the mean map layer performs: $D$ scaled $1 \times 1$ 
convolutions that are biased by uniformly random offsets, a cosine nonlinearity layer, and global-average pooling. One can see that $\hat{\mu}_{\Xcal_i}$ \eqref{eq:hatmu} is computed as the mean of the set of vectors $\left\{ z(C_{i,j,l})\,|\, 1 \leq j \leq h,\ 1 \leq l \leq w \right\}$. Let $Z_i \in \R^{D \times h \times w}$ be the tensor where 
\begin{align}
Z_i(d,j,l) = [z(C_{i,j,l})]_d = \cos\left( \omega_d^TC_{i,j,l} + b_d  \right) \label{eq:convrf},
\end{align}
and $\{b_d\}_{d=1}^D \iid \unif[0,2\pi]$ 
are drawn before training time and are held fixed throughout. 
Note that we have dropped the constant $\tfrac{1}{\sqrt{2D}}$ since it may be factored into linear operations above in the network. To compute these vectors, and their mean we perform (see Figure \ref{fig:dmm}):
\begin{description}
  \item[Frequency Convolution] The term $\omega_d^TC_{i,j,l} + b_d$ in \eqref{eq:convrf} is just a $1 \times 1$ convolution. Hence, we perform $D$ $1 \times 1$ convolutions (one for each frequency $\omega_d$).
  \item[Cosine] $Z_i$ is computed via a cosine layer that computes element-wise cosines to the convolutions above.
  \item[Global Pooling] The computation of the $\ith{d}$ element of $\hat{\mu}_{\Xcal_i}$ is just the average:
\begin{align}
[\hat{\mu}_{\Xcal_i}]_d = \frac{1}{hw} \sum_{j=1}^h \sum_{l=1}^w Z_i(d,j,l).
\end{align}
I.e. $\hat{\mu}_{\Xcal_i}$ is computed via global average pooling of the channels in $Z_i$. 
\end{description}
Hence, it is clear that the mean map layer is efficient to implement and train in terms of existing CNN operations and layers.

Note that in practice one may learn the kernel used in the mean map by learning frequencies $\{\omega_d\}_{d=1}^D$, or one may consider a fixed kernel by drawing $\{\omega_d\}_{d=1}^D \iid \rho$ (where $\rho$ is the spectral density for the kernel, e.g. $\N(0, \sigma^{-2} I)$ for the RBF kernel). Furthermore, one may learn the scale for a fixed kernel; e.g. one may learn the bandwidth of the RBF kernel in the MME by drawing $\{\omega_d\}_{d=1}^D  \iid \N(0, I)$ and parameterizing the cosine layer in terms of $a\in\R$ as:
\begin{align}
Z_i(d,j,l) = \cos\left( e^a \omega_d^TC_{i,j,l} + b_d  \right) \label{eq:cosscale},
\end{align}
where here $\sigma^2 = e^{2a}$, and can be learned via back-propagation.

After being fed through the MME layer (with the global-average pooling of random features), the $\ith{i}$ image in a batch will have the mean map embedding of its top-level convolution features, $\hat\mu_{\Xcal_i} \in \R^D$, computed. These vectors can then then be fed through the network in a fully connected fashion to perform learning w.r.t. a loss. Note that linear operations of the MME vectors, $\hat\mu_{\Xcal_i}^T \psi$ where $\psi \in \R^D$, may be interpreted as an RKHS inner-product with an element $f \in \Hcal$  \eqref{eq:rksdot}. 
Moreover, it is straightforward to perform back-propagation through the MME layer 
by calculating derivatives of (\ref{eq:convrf}, \ref{eq:cosscale}) w.r.t. inputs and parameters.

\section{Empirical Evaluation} \label{sec:experiments}

We demonstrate the effectiveness of our approach on two image classification problems.
All experiments are implemented in the Caffe deep learning framework \cite{caffe};
code will be made publicly available.

\subsection{Synthetic Data}
\begin{figure*}[t]
\centering
\bgroup
\setlength{\tabcolsep}{1pt}
\tiny
\begin{tabular}{cccccccc}
\includegraphics[width=.09\textwidth]{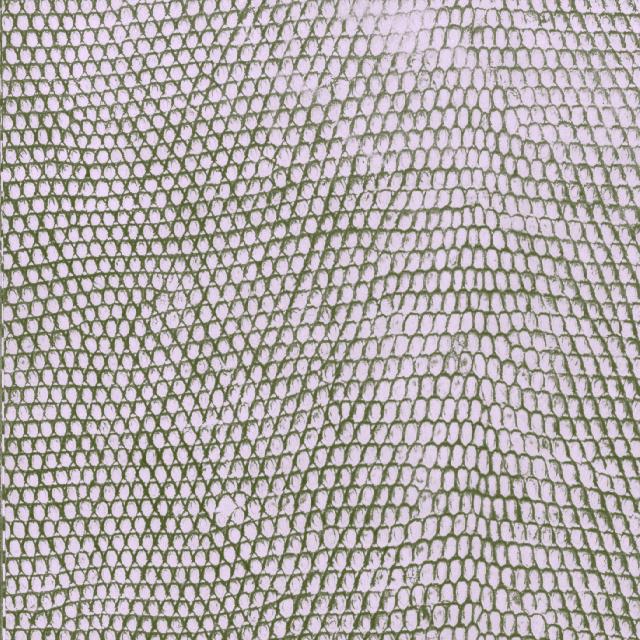} &
\includegraphics[width=.09\textwidth]{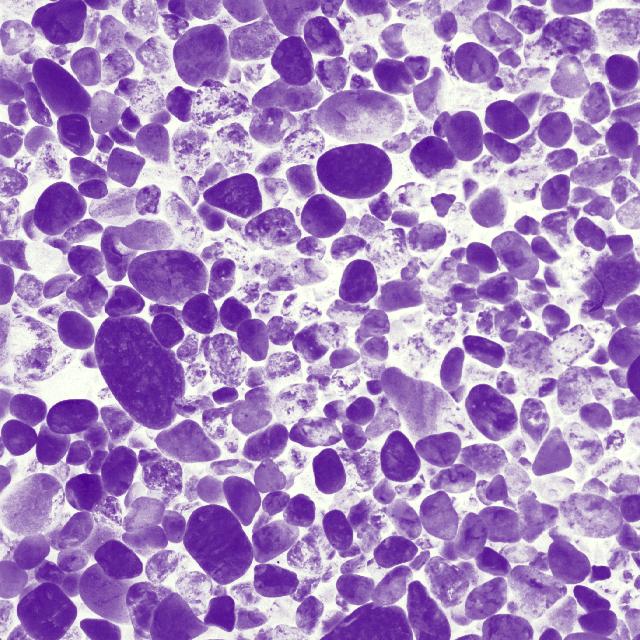} &
\includegraphics[width=.09\textwidth]{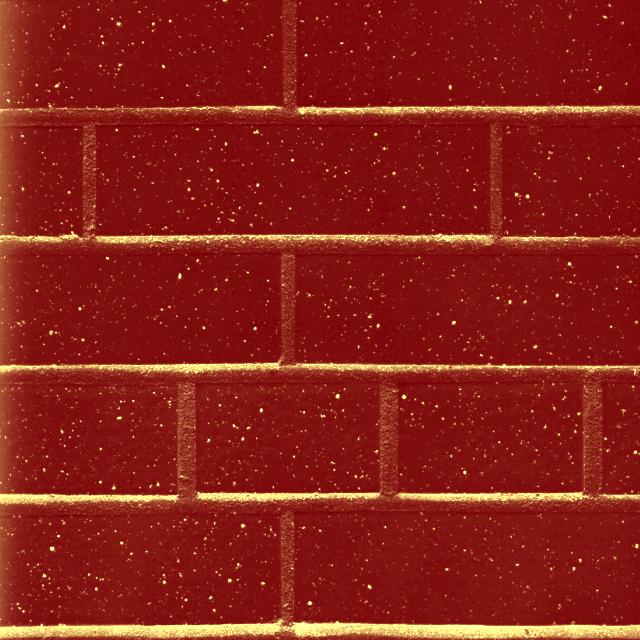} &
\includegraphics[width=.09\textwidth]{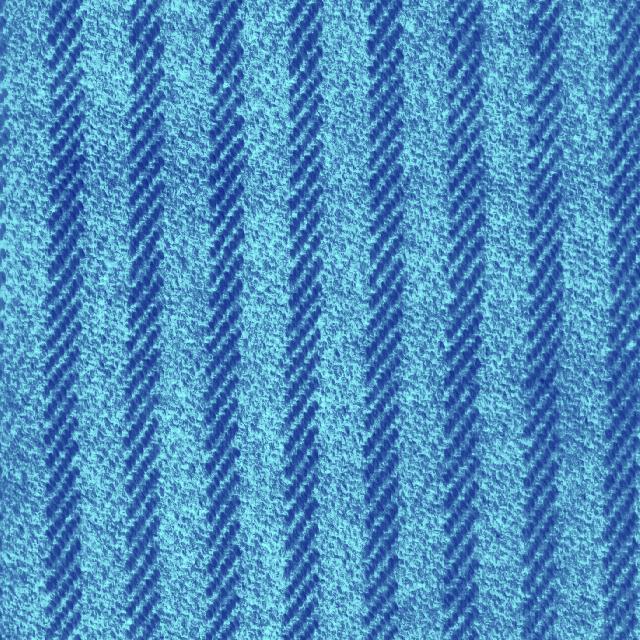} &
\includegraphics[width=.09\textwidth]{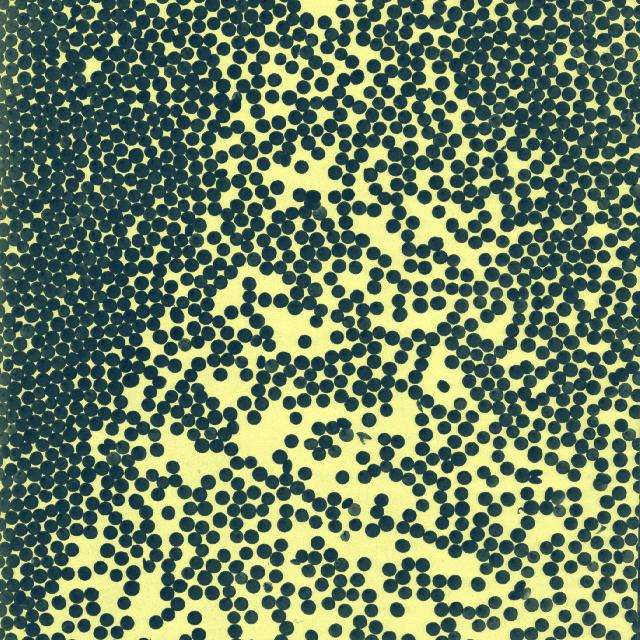} &
\includegraphics[width=.09\textwidth]{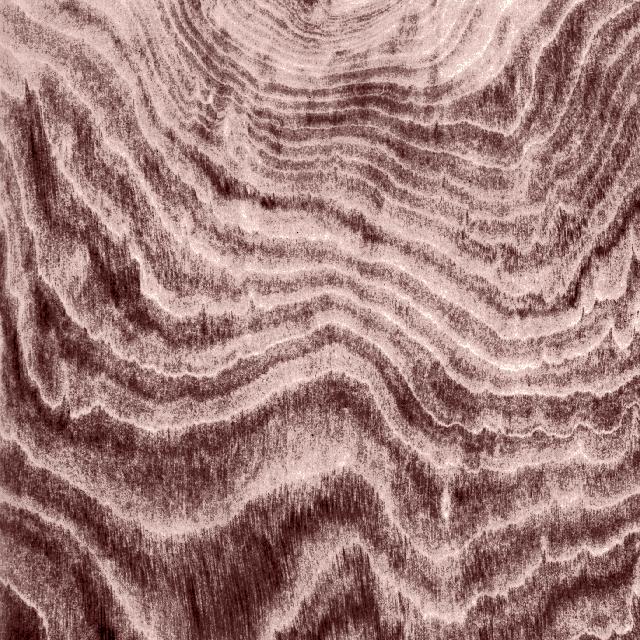} &
\includegraphics[width=.09\textwidth]{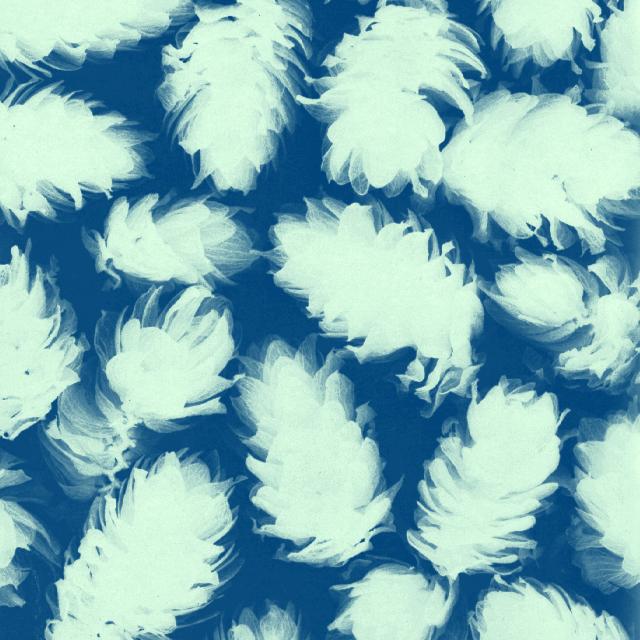} &
\includegraphics[width=.09\textwidth]{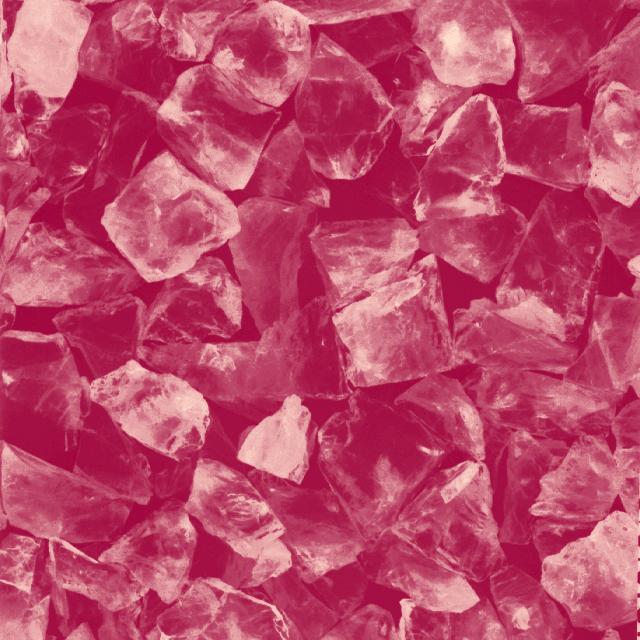}
\\
example texture a &
example texture b &
example texture c &
example texture d &
example texture e &
example texture f &
example texture g &
example texture h
\\
\includegraphics[width=.09\textwidth]{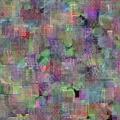} &
\includegraphics[width=.09\textwidth]{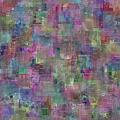} &
\includegraphics[width=.09\textwidth]{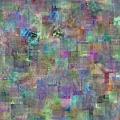} &
\includegraphics[width=.09\textwidth]{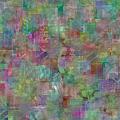} &
\includegraphics[width=.09\textwidth]{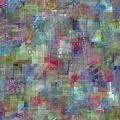} &
\includegraphics[width=.09\textwidth]{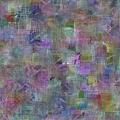} &
\includegraphics[width=.09\textwidth]{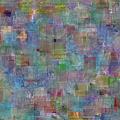} &
\includegraphics[width=.09\textwidth]{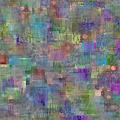}
\\
class 1 &
class 2 &
class 3 &
class 4 &
class 5 &
class 6 &
class 7 &
class 8
\end{tabular}
\egroup
\vspace{1 mm}
\caption{Top row: example images from the CBT database. Bottom row: random synthetic data images from the 8 classes.}
\label{fig:syn-samples}
\end{figure*}

Below we show the ability of the mean map layer to uncover visual distributional patterns with a simple synthetic data example. 

We generate data using a bank of $M=112$ texture images found in the Colored Brodatz Texture (CBT) database \cite{textbrodatz}. Each class represents a distribution of patches in the generated images over the texture bank. We consider a dataset of 8 such classes, generated at random.
Specifically, the $\ith{i}$ class corresponds to a mixture of 30 Dirichlet distributions, $P_i$, where each mixture component has support in the $M$-dimensional simplex (one dimension for each texture image). Given the Dirichlet mixture for the class, we generate a $120 \times 120$ image as follows: pick a random location for a $12 \times 12$ patch uniformly at random, draw $\pi\sim P_i$, pick a random patch from each image $\{p_j\}_{j=1}^M$; set the patch to $p = \sum_{j=1}^M \pi_j p_j$, overlay on chosen location, repeat $1500$ times. We generated a set of $800$ total training images ($100$ per class) and $4000$ test images ($500$ per-class).

We used a simple CNN network with the following structure: a convolution of 58 filters with kernel sized $12 \times 12$ and a stride of 1;  a ReLU layer; $12 \times 12$ max pooling with a stride of 6. We appended a mean map layer to the end of the network (\texttt{mml}). We also compared to appending two hidden layers each of $4096$ nodes with ReLU non-linearity (\texttt{hid}), or directly classifying on the max pooled features (\texttt{lin}). Each network was trained with SGD.

\begin{figure}[h]
  \centering
  \includegraphics[width=.4\textwidth]{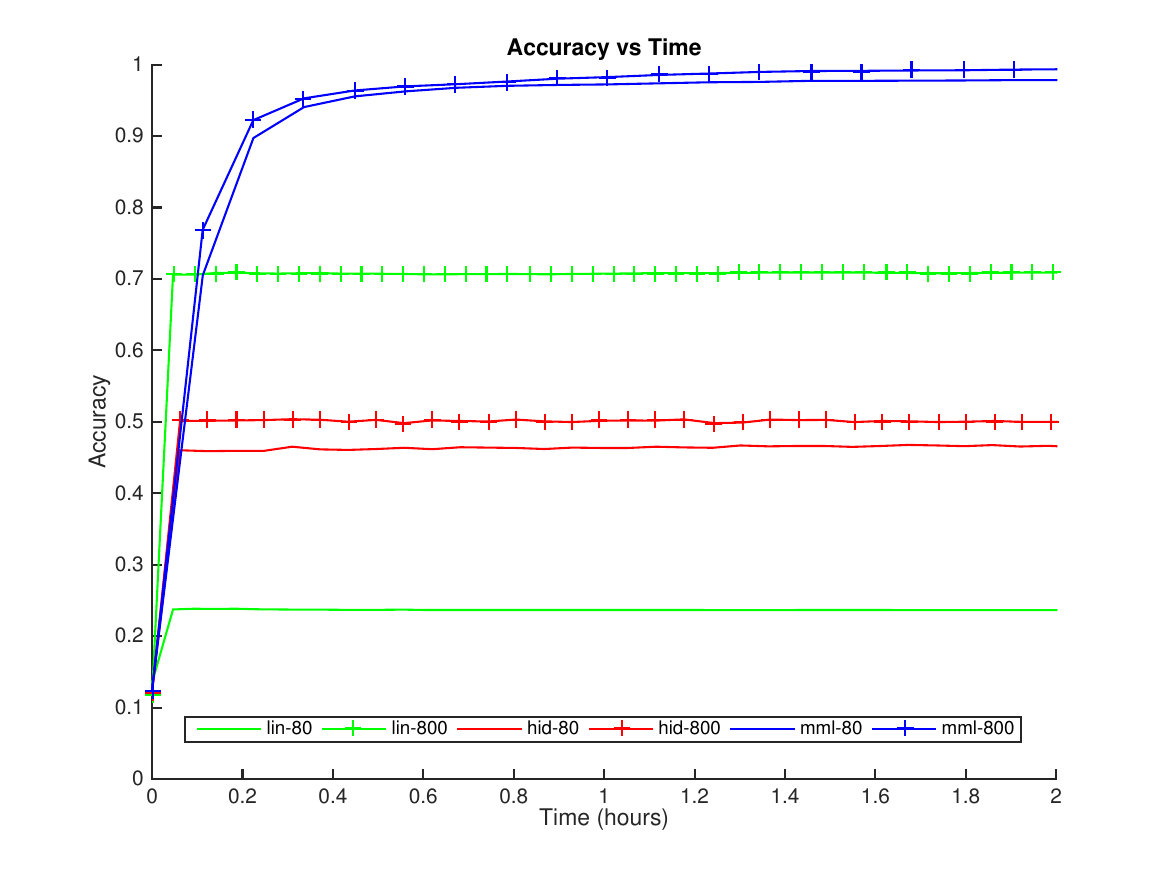}
  \caption{Accuracy vs. time for synthetic data experiment. Blue curves correspond to the accuracies for the \texttt{mml} network with the mean map layer. Green curves correspond to the \texttt{lin} network, which feeds max-pooled filters directly to a classifier. Lastly, red curves correspond to the \texttt{hid} network, which uses two hidden layers to classify. Solid and crossed lines indicate a training set of $80$ and $800$ respectively. \label{fig:synaccu}}
\end{figure}

We look at the test accuracies of the aforementioned three networks on a subset of $80$ training images ($10$ from each class) and all $800$ (see Figure \ref{fig:synaccu}). It is interesting to note that even with only $10$ instances from each class \texttt{mml} is able to almost perfectly distinguish the classes, and outperforms the other networks by a large margin. This is because the \texttt{mml} has the built in mechanisms to learn over distributions of images features and thus may do so in a sample efficient fashion. 

\begin{figure*}[t]
\centering
\bgroup
\setlength{\tabcolsep}{1pt}
\tiny
\begin{tabular}{cccccccccc}
\includegraphics[width=.09\textwidth,height=.09\textwidth]{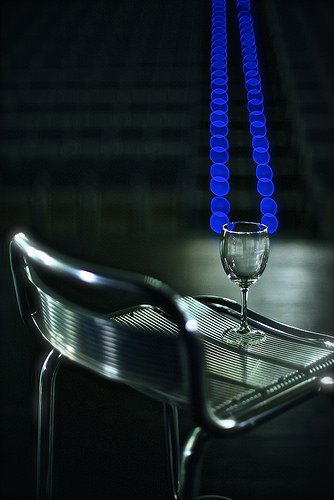} &
\includegraphics[width=.09\textwidth,height=.09\textwidth]{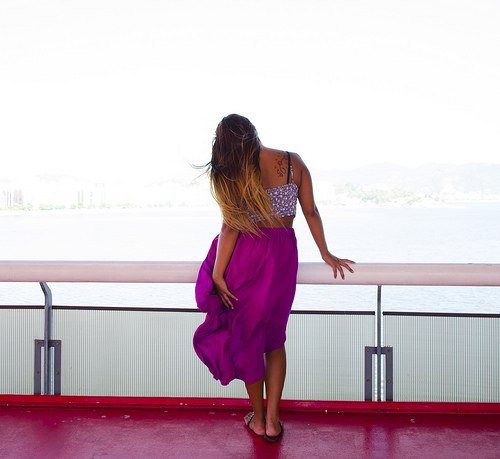} &
\includegraphics[width=.09\textwidth,height=.09\textwidth]{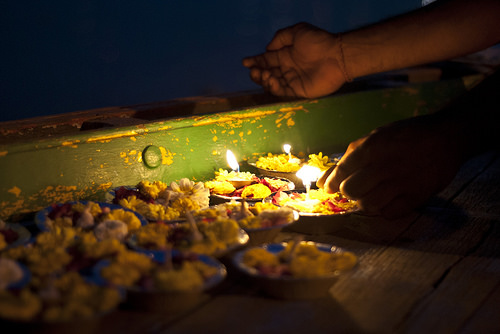} &
\includegraphics[width=.09\textwidth,height=.09\textwidth]{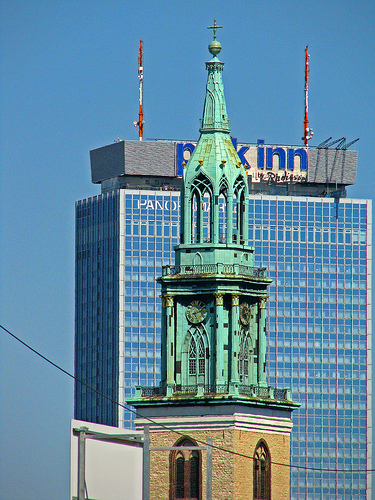} &
\includegraphics[width=.09\textwidth,height=.09\textwidth]{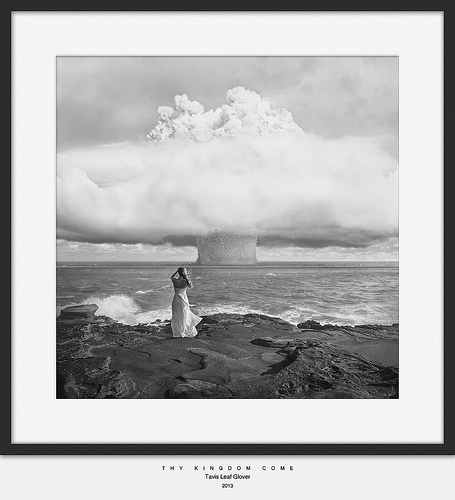} &
\includegraphics[width=.09\textwidth,height=.09\textwidth]{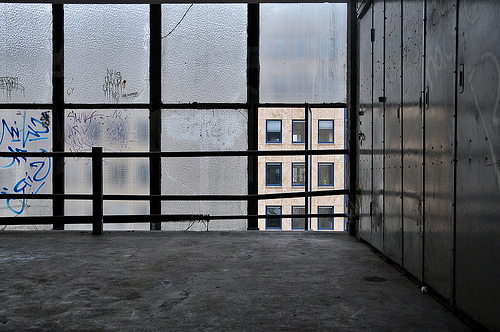} &
\includegraphics[width=.09\textwidth,height=.09\textwidth]{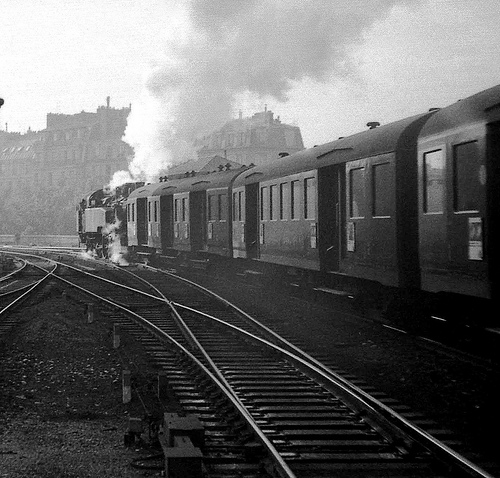} &
\includegraphics[width=.09\textwidth,height=.09\textwidth]{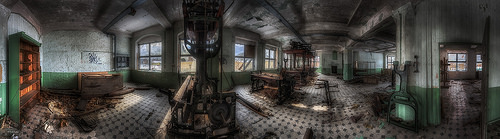} &
\includegraphics[width=.09\textwidth,height=.09\textwidth]{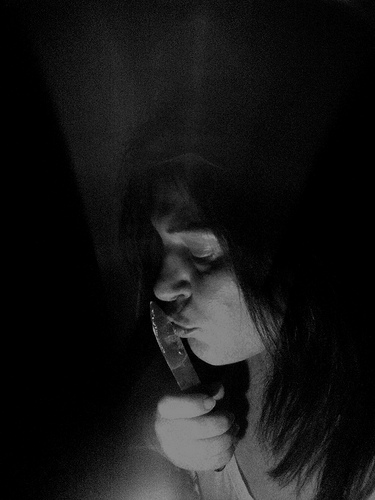} &
\includegraphics[width=.09\textwidth,height=.09\textwidth]{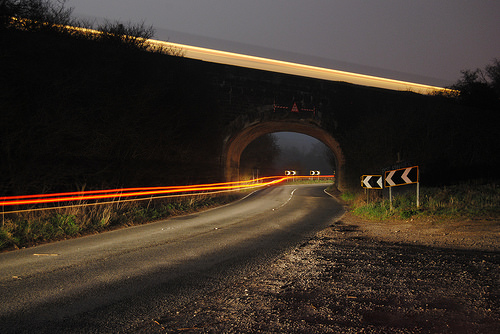} \\
Bokeh &
Bright &
Depth of Field &
Detailed &
Ethereal &
Geometric &
Hazy &
HDR &
Horror &
Long Exposure \\
\includegraphics[width=.09\textwidth,height=.09\textwidth]{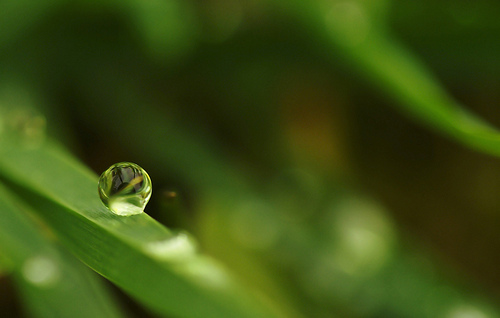} &
\includegraphics[width=.09\textwidth,height=.09\textwidth]{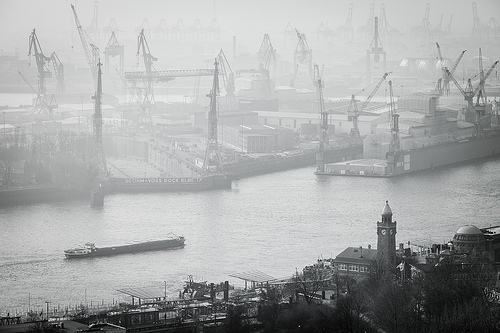} &
\includegraphics[width=.09\textwidth,height=.09\textwidth]{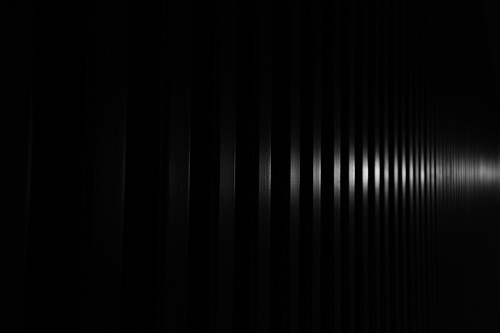} &
\includegraphics[width=.09\textwidth,height=.09\textwidth]{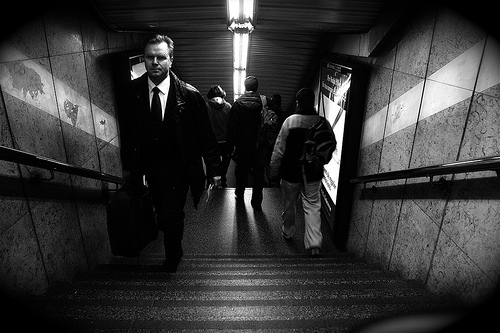} &
\includegraphics[width=.09\textwidth,height=.09\textwidth]{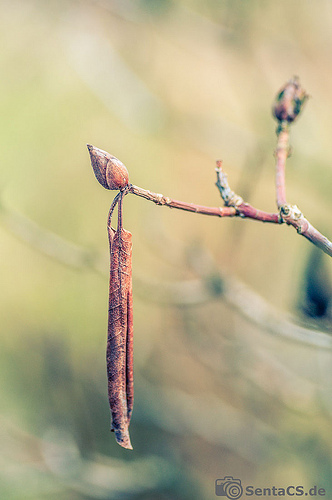} &
\includegraphics[width=.09\textwidth,height=.09\textwidth]{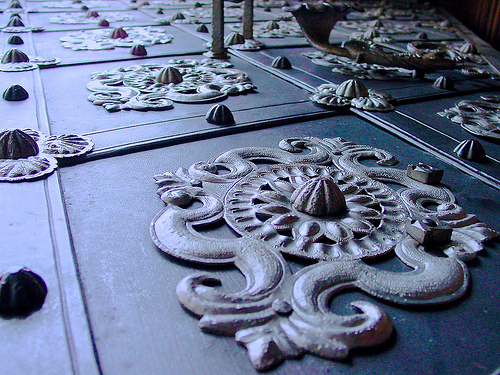} &
\includegraphics[width=.09\textwidth,height=.09\textwidth]{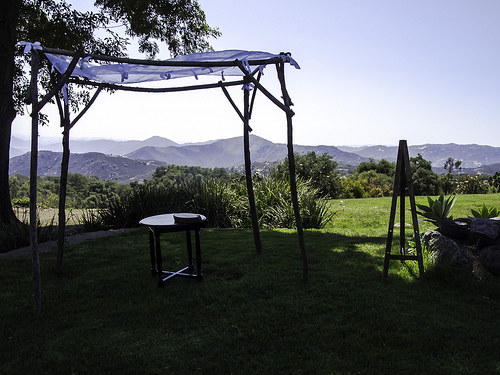} &
\includegraphics[width=.09\textwidth,height=.09\textwidth]{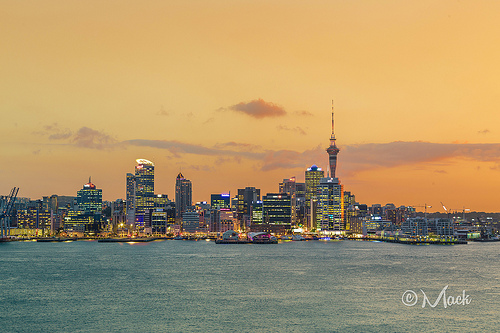} &
\includegraphics[width=.09\textwidth,height=.09\textwidth]{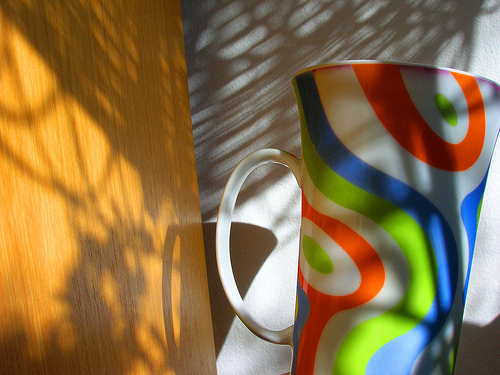} &
\includegraphics[width=.09\textwidth,height=.09\textwidth]{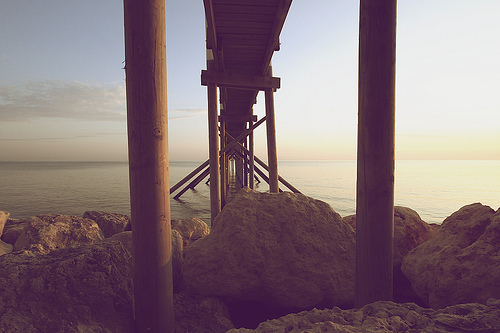} \\
Macro &
Melancholy &
Minimal &
Noir &
Pastel &
Romantic &
Serene &
Sunny &
Texture &
Vintage
\end{tabular}
\egroup
\vspace{1 mm}
\caption{A random image from each of the 20 classes in Flickr Style.}
\label{fig:flickr-samples}
\end{figure*}

\begin{figure*}[t]
\centering
\bgroup
\setlength{\tabcolsep}{1pt}
\tiny
\begin{tabular}{ccccccccccccc}
\includegraphics[width=.06\textwidth,height=.07\textwidth]{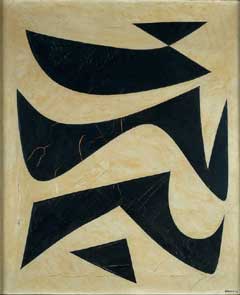} &
\includegraphics[width=.06\textwidth,height=.07\textwidth]{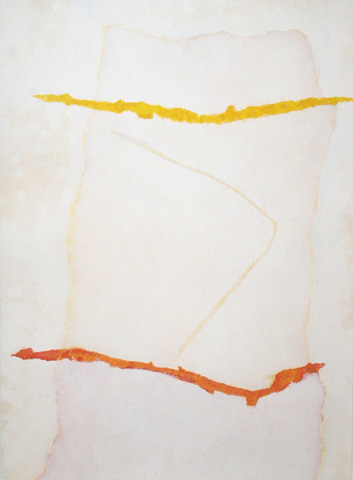} &
\includegraphics[width=.06\textwidth,height=.07\textwidth]{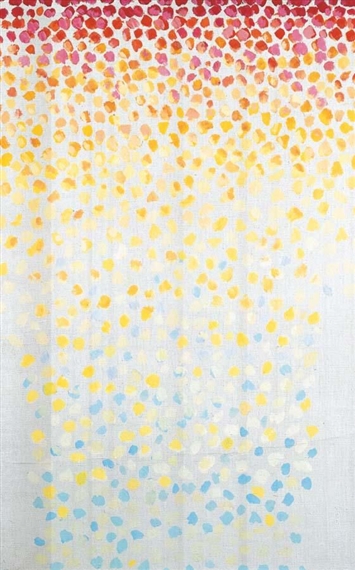} &
\includegraphics[width=.06\textwidth,height=.07\textwidth]{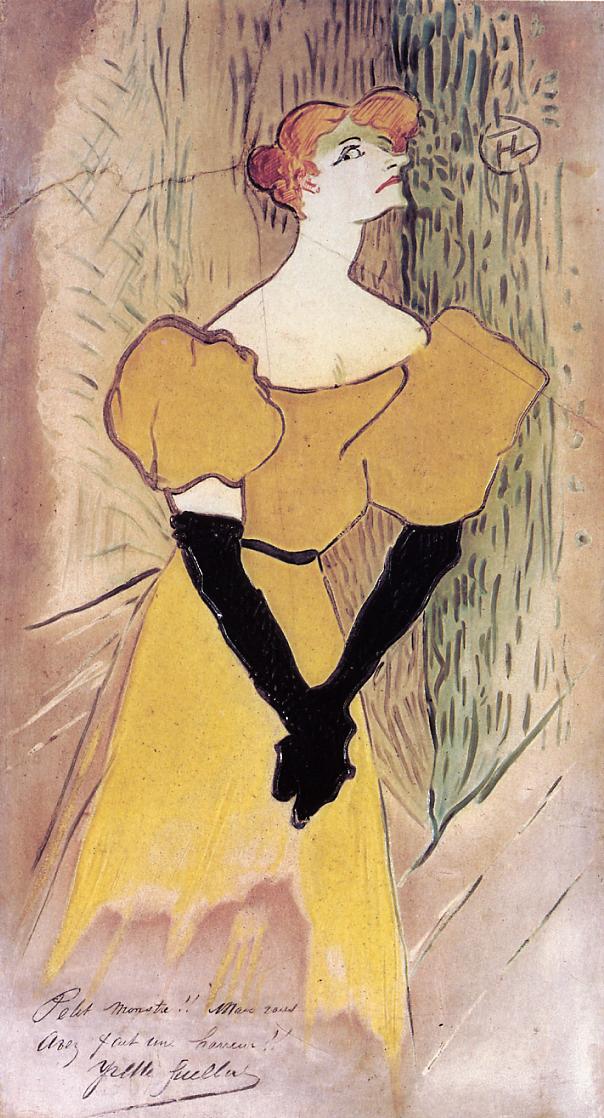} &
\includegraphics[width=.06\textwidth,height=.07\textwidth]{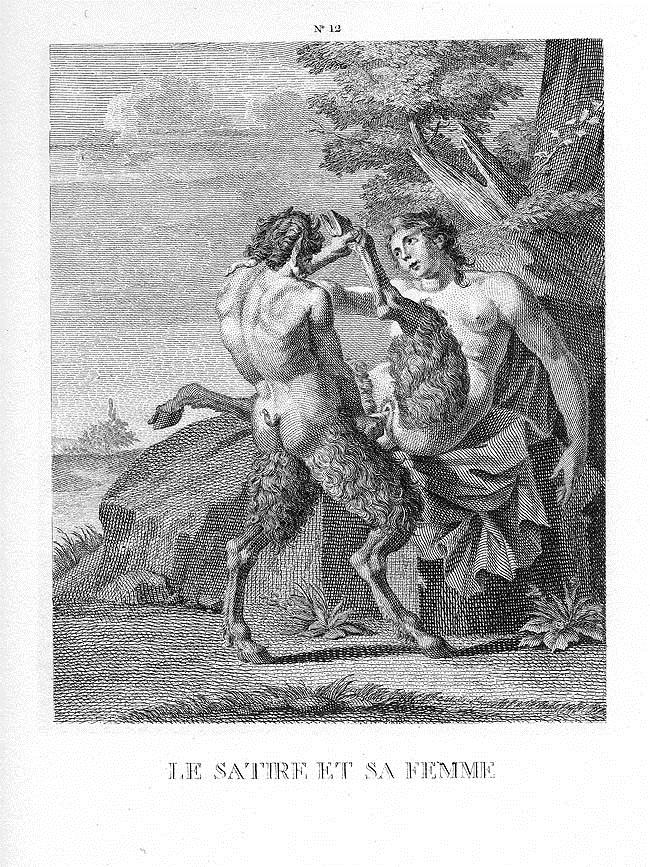} &
\includegraphics[width=.06\textwidth,height=.07\textwidth]{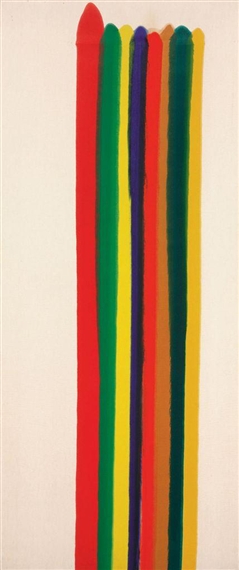} &
\includegraphics[width=.06\textwidth,height=.07\textwidth]{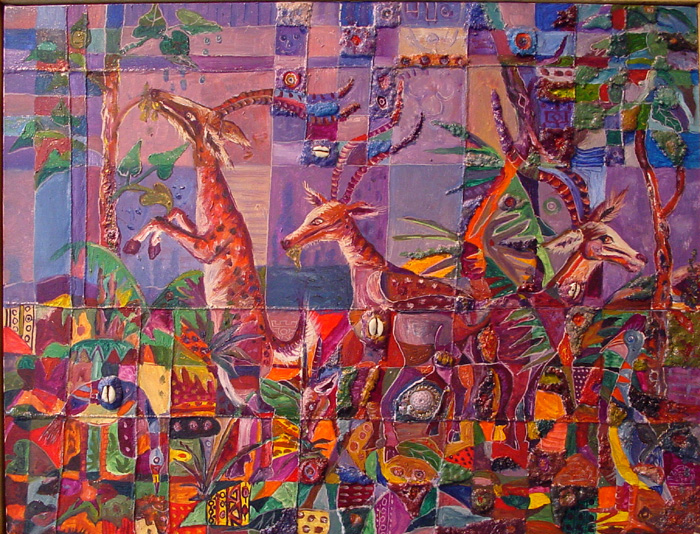} &
\includegraphics[width=.06\textwidth,height=.07\textwidth]{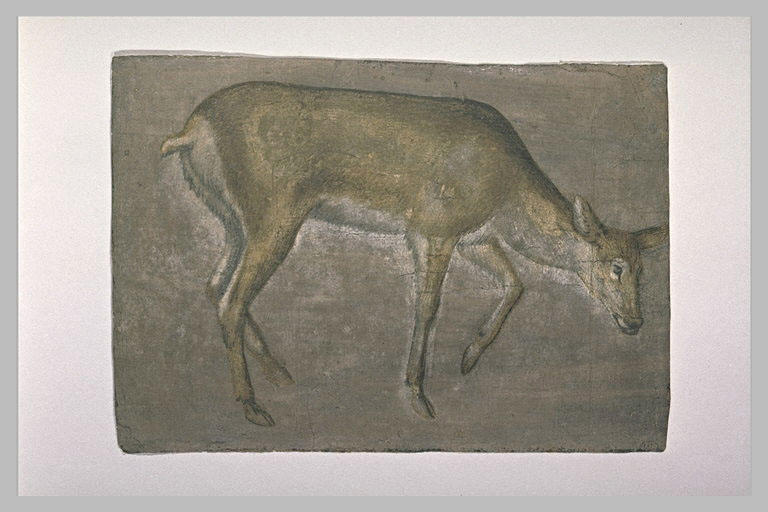} &
\includegraphics[width=.06\textwidth,height=.07\textwidth]{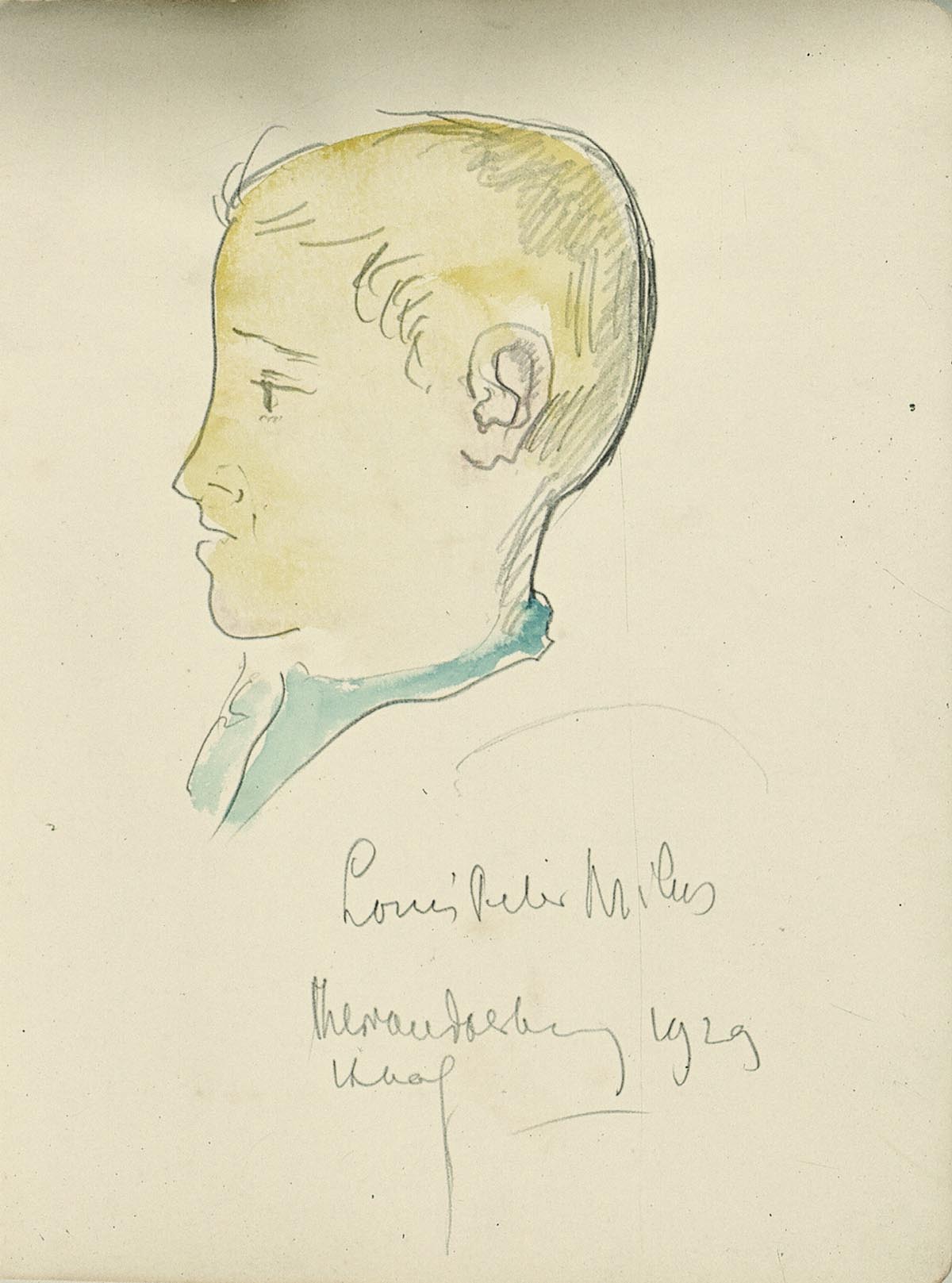} &
\includegraphics[width=.06\textwidth,height=.07\textwidth]{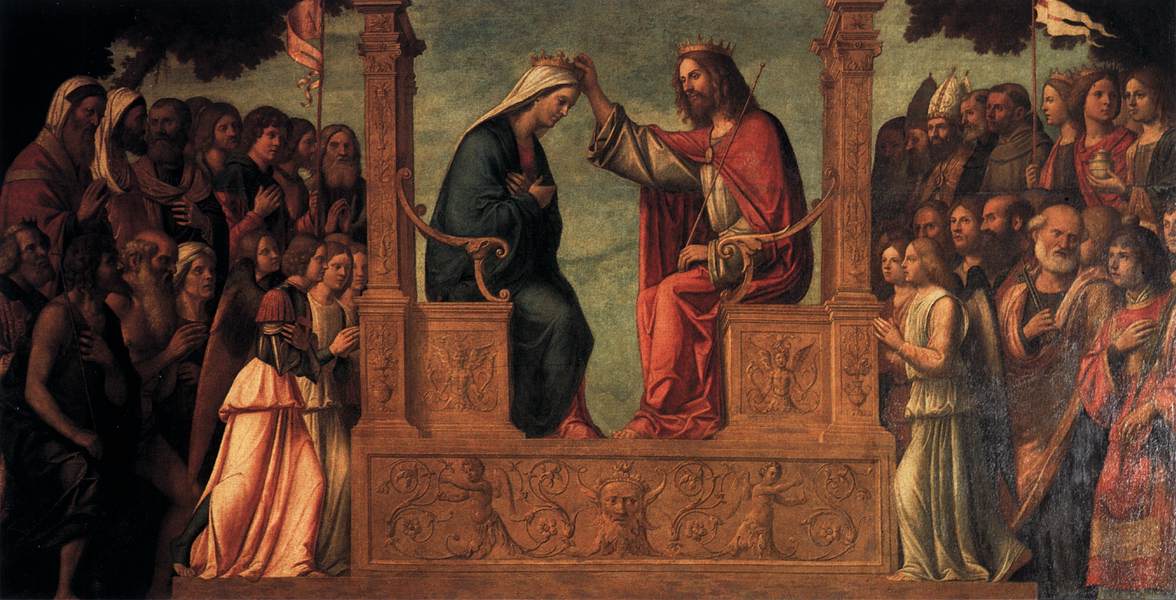} &
\includegraphics[width=.06\textwidth,height=.07\textwidth]{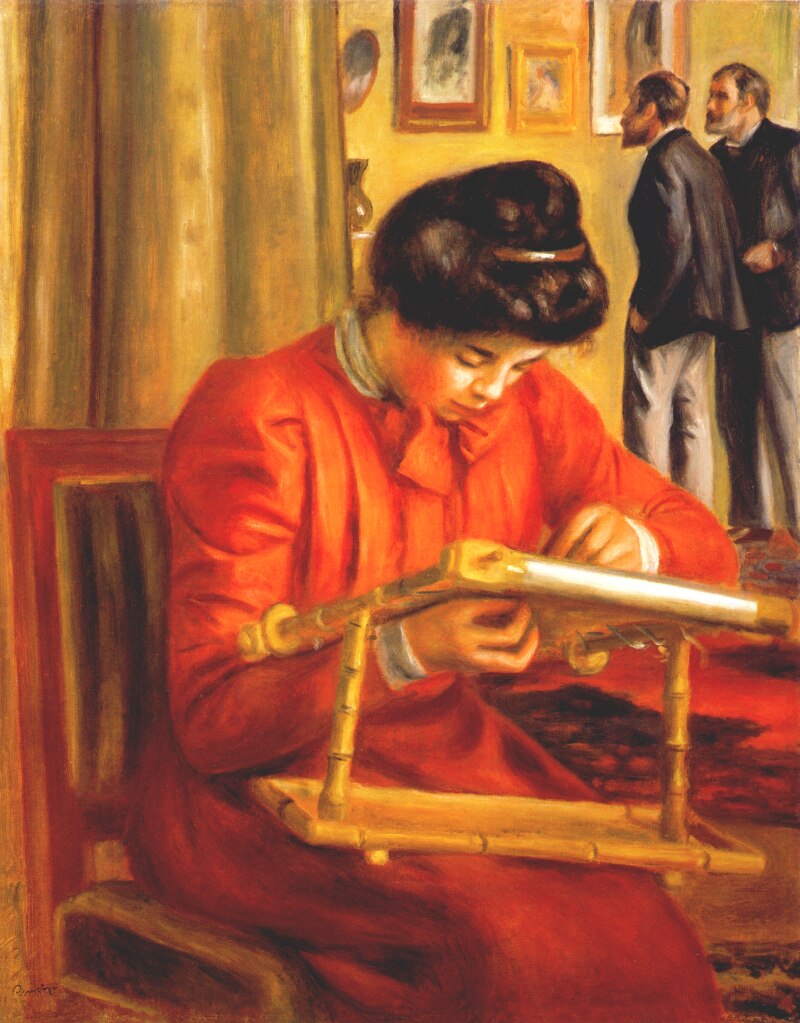} &
\includegraphics[width=.06\textwidth,height=.07\textwidth]{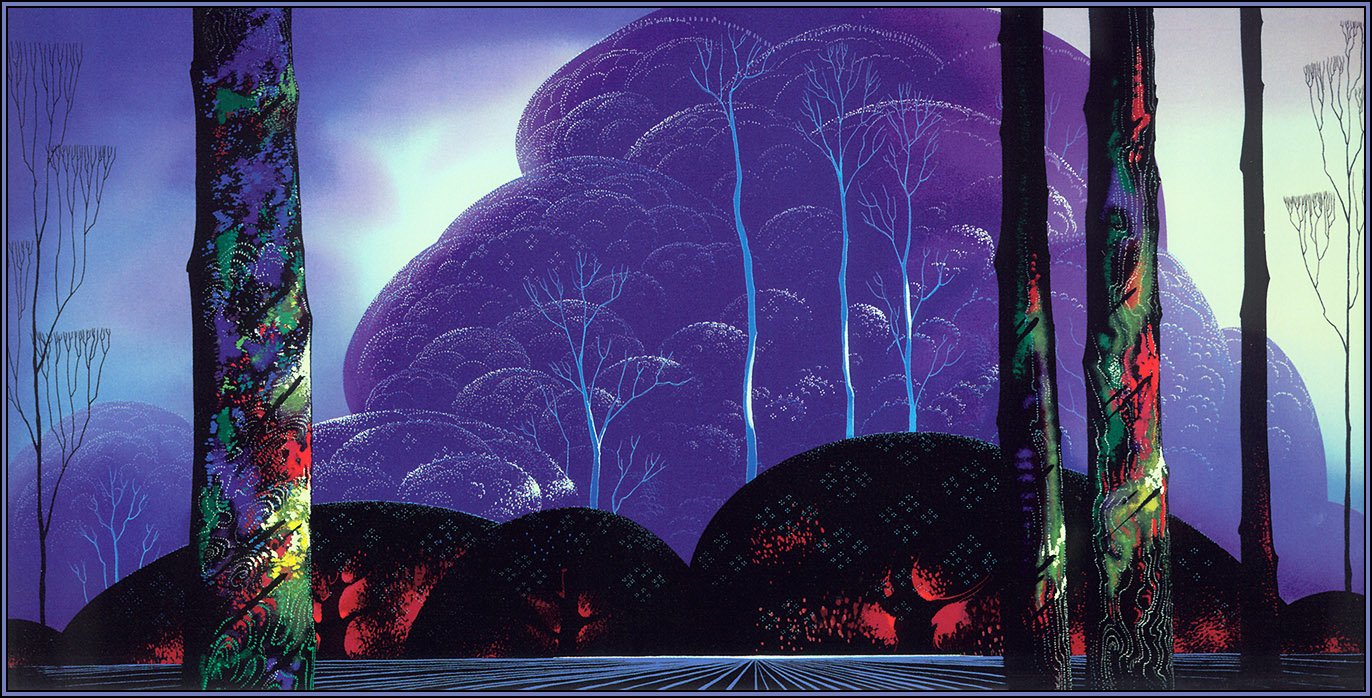} &
\includegraphics[width=.06\textwidth,height=.07\textwidth]{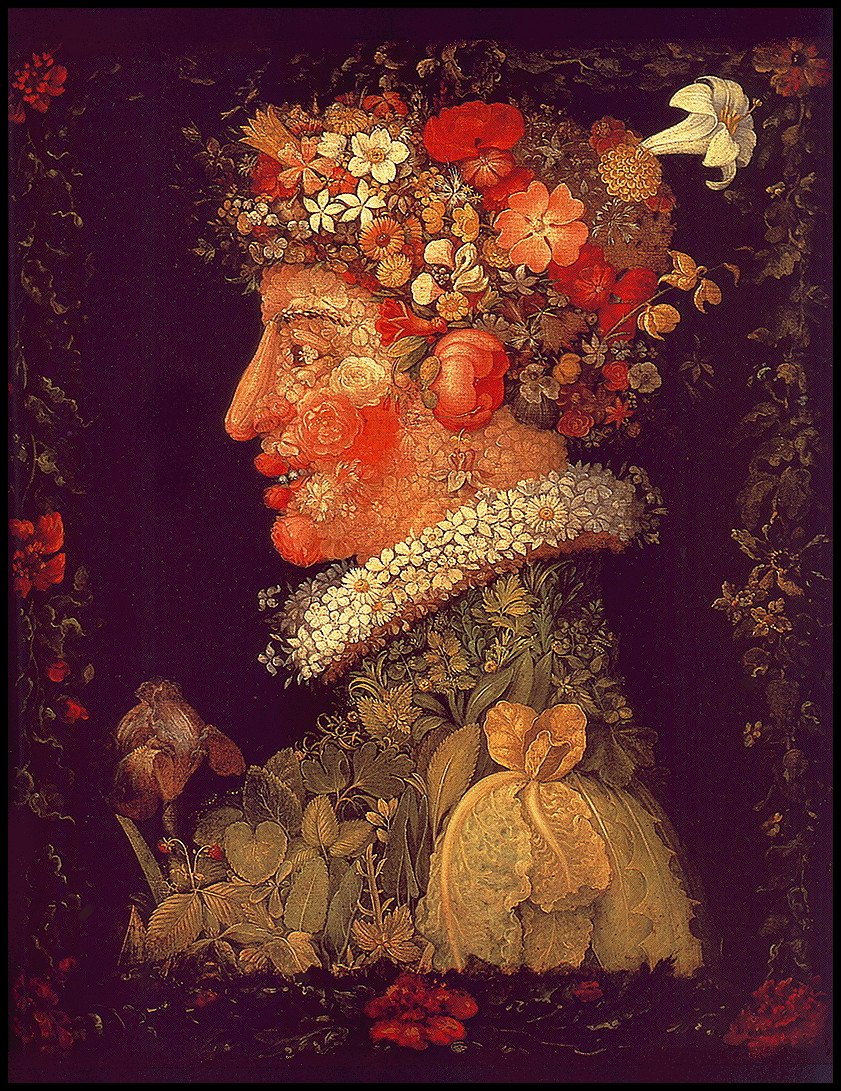} \\
Abstract Art &
Abstr.\ Express\. &
Art Informel &
Art Nouveau &
Baroque &
Color Field Painting &
Cubism &
Early Renaissance &
Expressionism &
High Renaissance &
Impressionism &
Magic Realism &
Mannerism \\
\includegraphics[width=.07\textwidth,height=.07\textwidth]{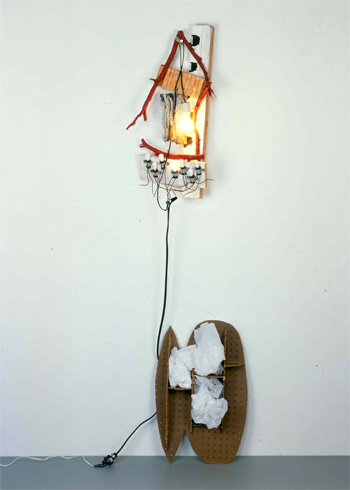} &
\includegraphics[width=.07\textwidth,height=.07\textwidth]{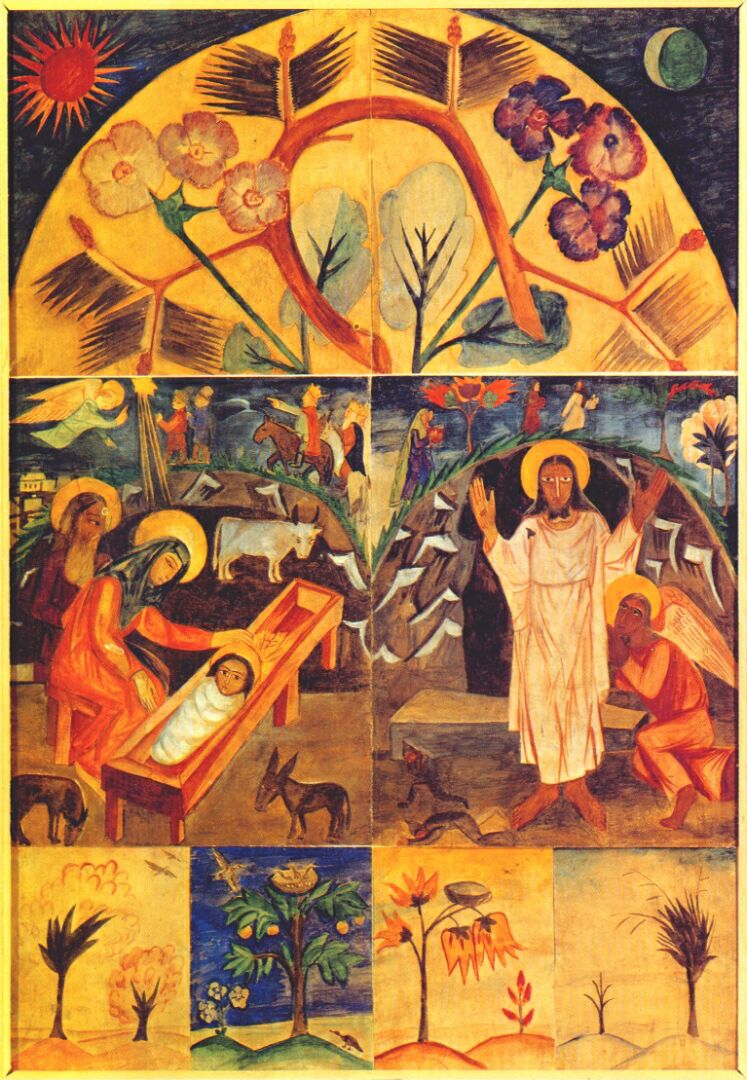} &
\includegraphics[width=.07\textwidth,height=.07\textwidth]{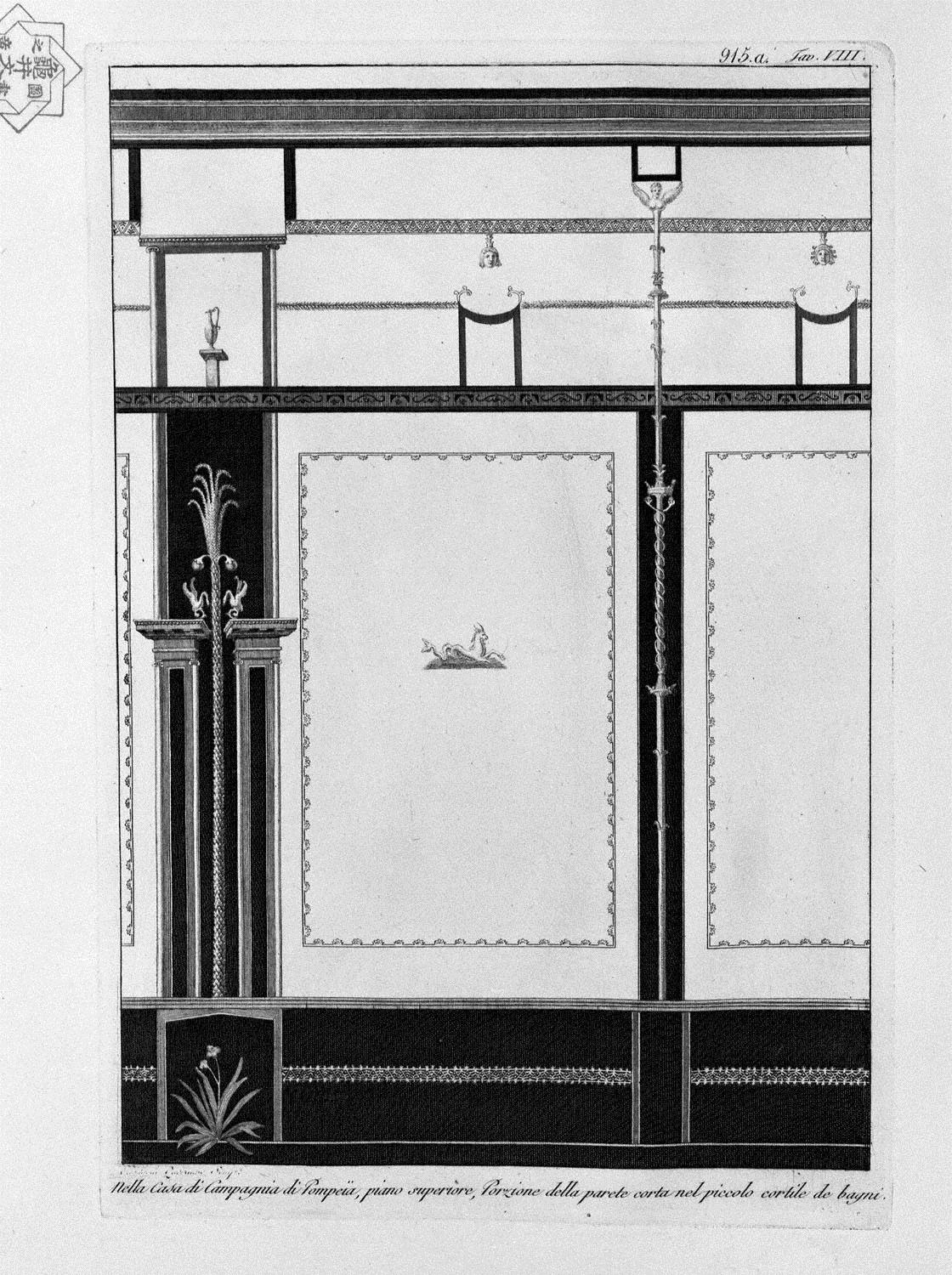} &
\includegraphics[width=.07\textwidth,height=.07\textwidth]{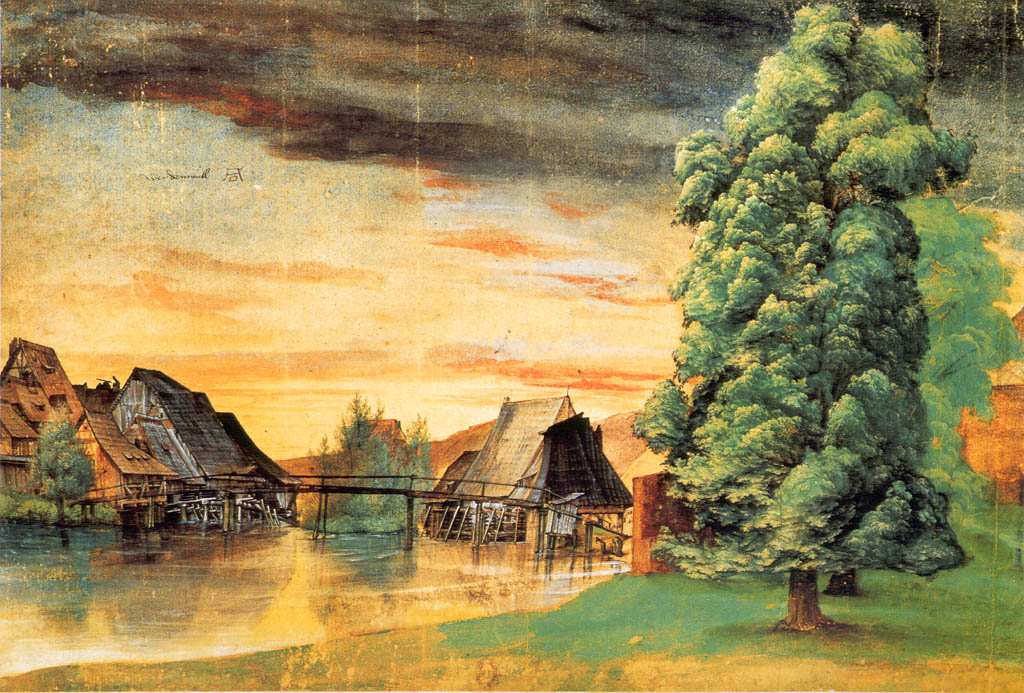} &
\includegraphics[width=.07\textwidth,height=.07\textwidth]{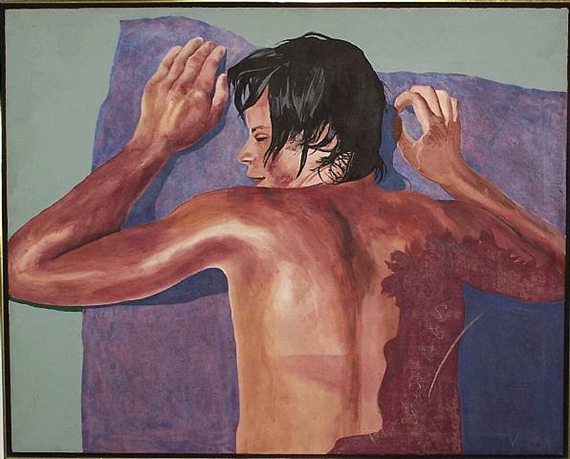} &
\includegraphics[width=.07\textwidth,height=.07\textwidth]{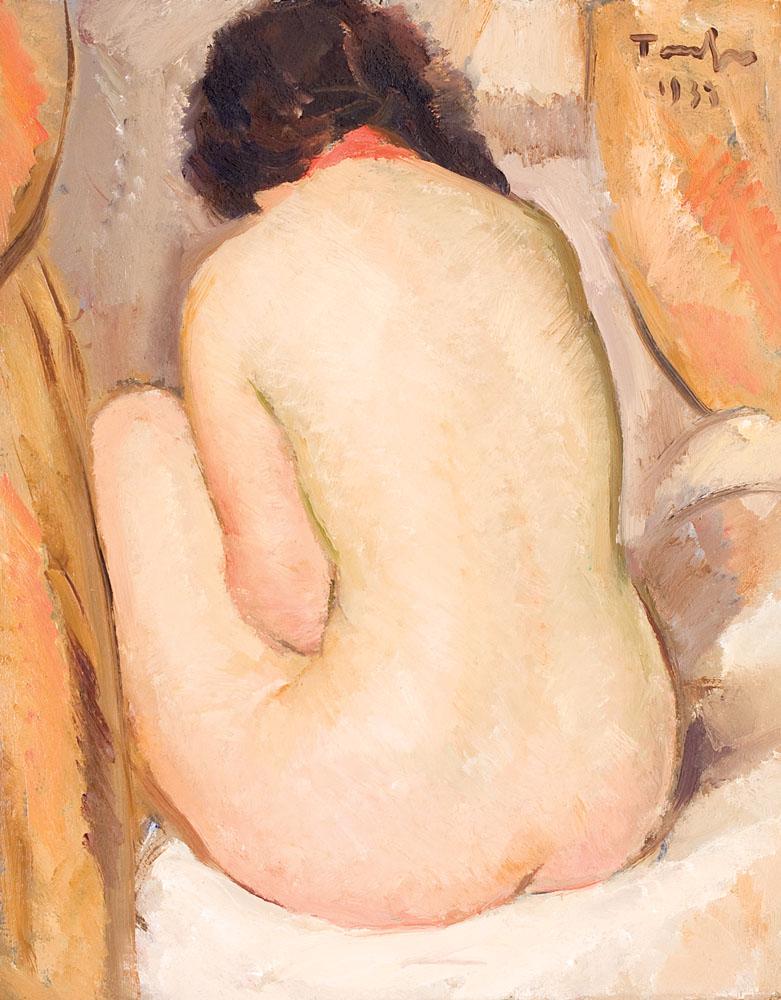} &
\includegraphics[width=.07\textwidth,height=.07\textwidth]{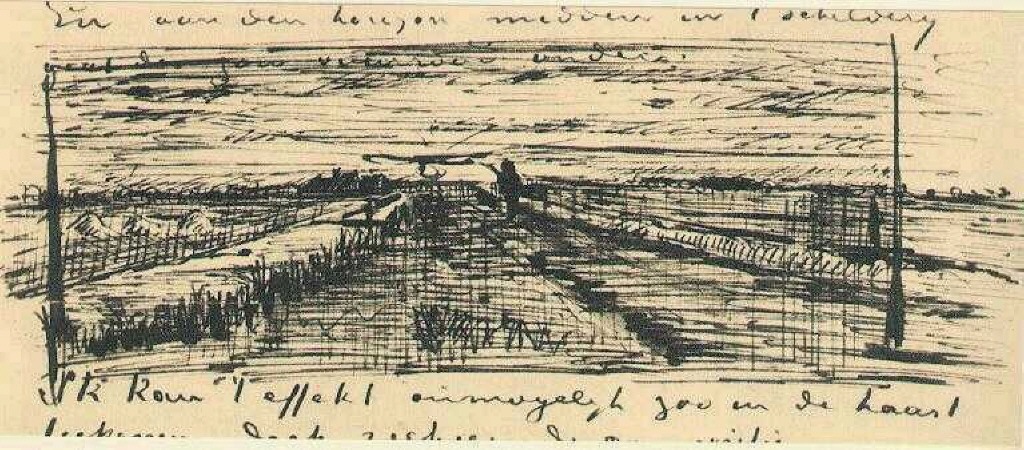} &
\includegraphics[width=.07\textwidth,height=.07\textwidth]{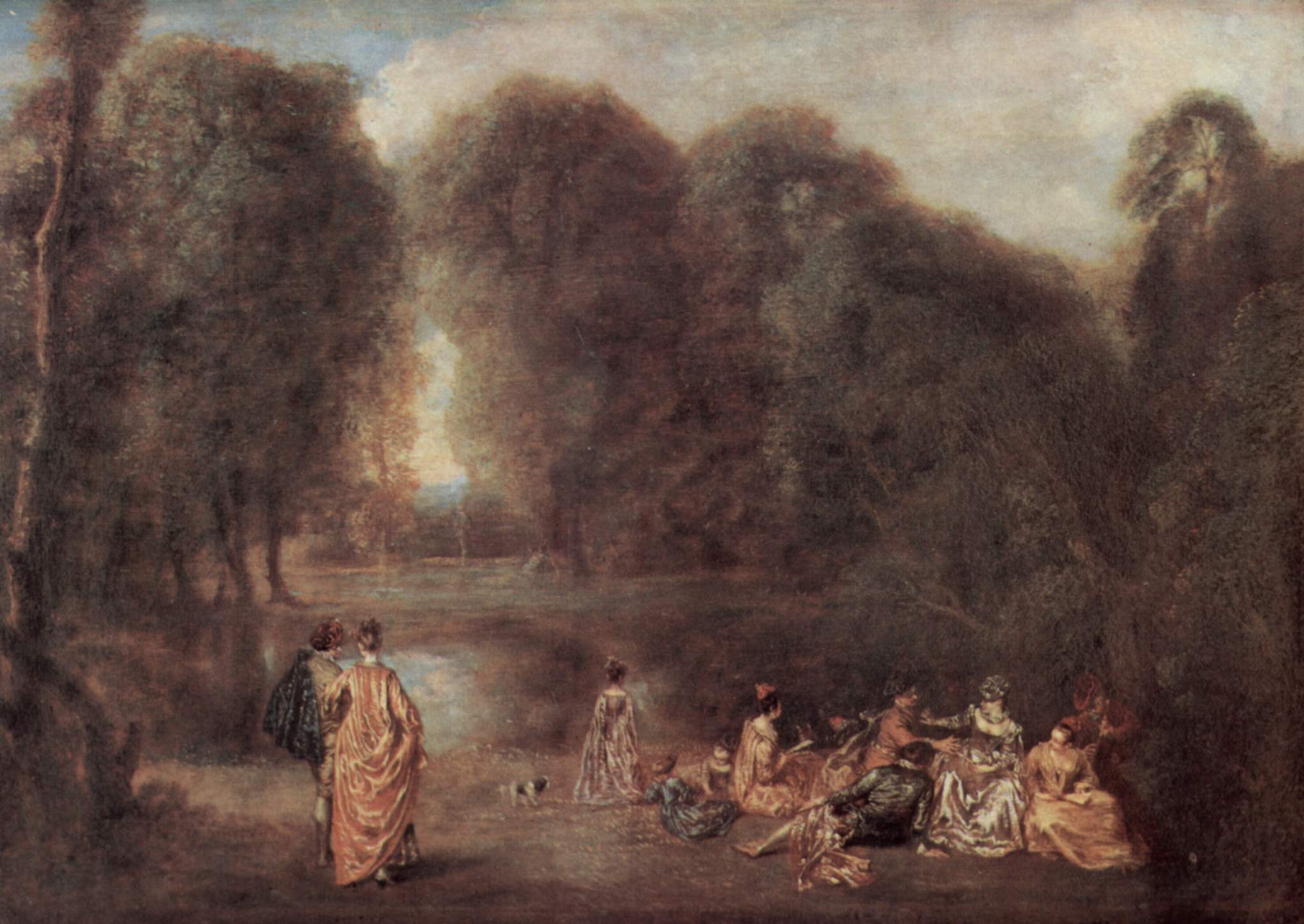} &
\includegraphics[width=.07\textwidth,height=.07\textwidth]{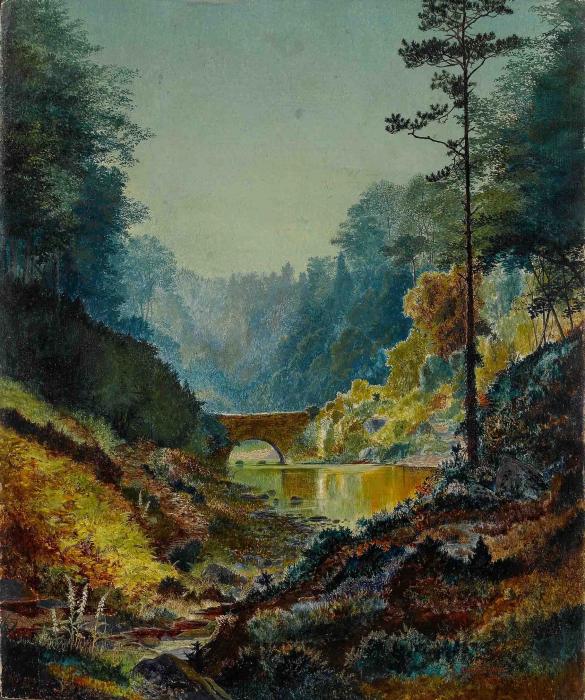} &
\includegraphics[width=.07\textwidth,height=.07\textwidth]{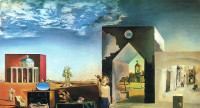} &
\includegraphics[width=.07\textwidth,height=.07\textwidth]{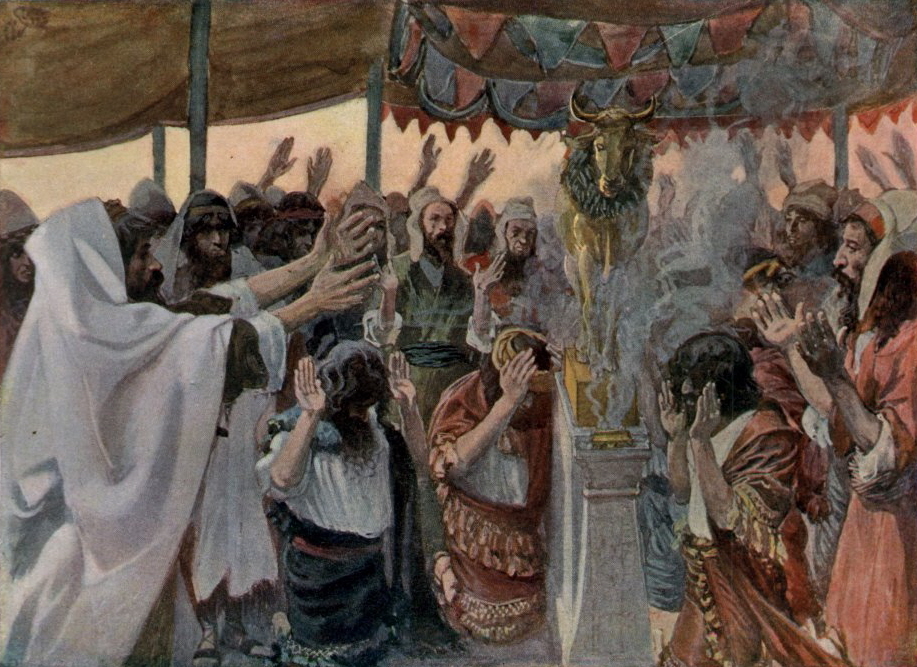} &
\includegraphics[width=.07\textwidth,height=.07\textwidth]{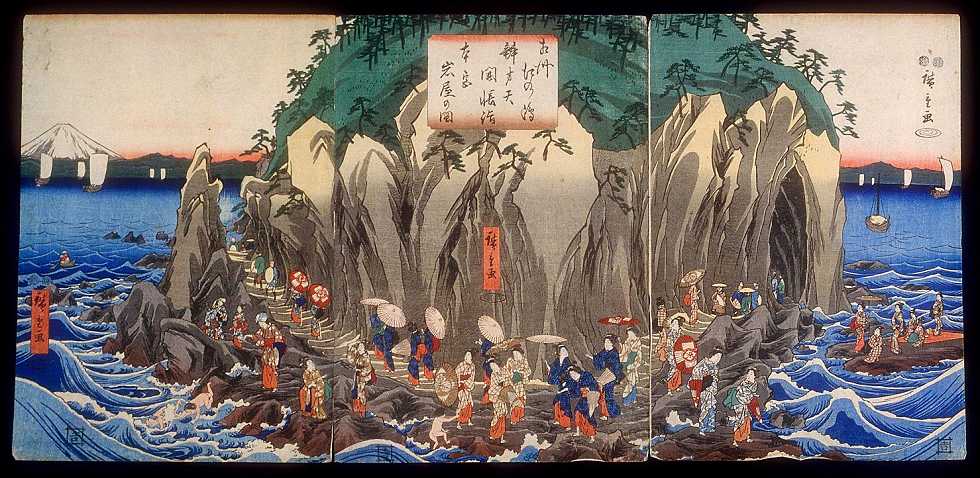} \\
Minimalism &
Naïve Art &
Neoclassicism &
Northern Renaissance &
Pop Art &
Post-Impressionism &
Realism &
Rococo &
Romanticism &
Surrealism &
Symbolism &
Ukiyo-e
\end{tabular}
\egroup
\vspace{1 mm}
\caption{One random image from each of the 25 classes in Wikipaintings.}
\label{fig:wikipaintings}
\end{figure*}

\begin{figure*}[t]
\centering
\bgroup
\setlength{\tabcolsep}{1pt}
\tiny
\begin{tabular}{cccccccccccc}
\includegraphics[width=.08\textwidth]{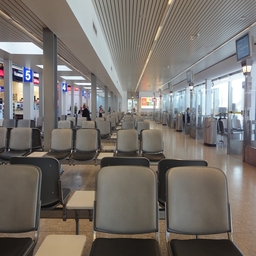} &
\includegraphics[width=.08\textwidth]{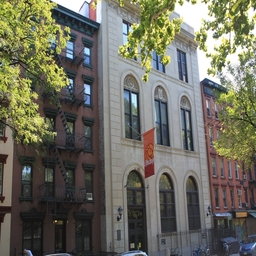} &
\includegraphics[width=.08\textwidth]{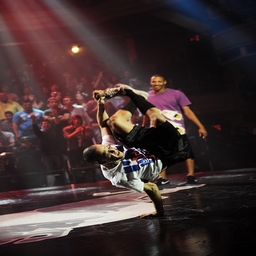} &
\includegraphics[width=.08\textwidth]{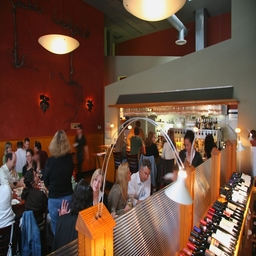} &
\includegraphics[width=.08\textwidth]{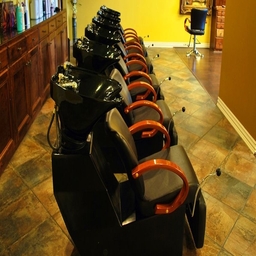} &
\includegraphics[width=.08\textwidth]{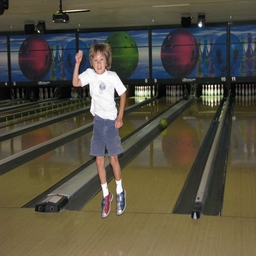} &
\includegraphics[width=.08\textwidth]{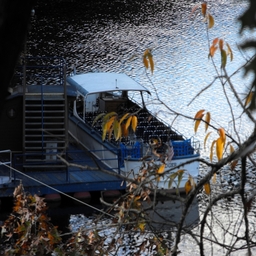} &
\includegraphics[width=.08\textwidth]{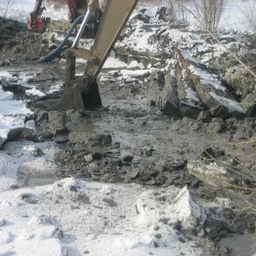} &
\includegraphics[width=.08\textwidth]{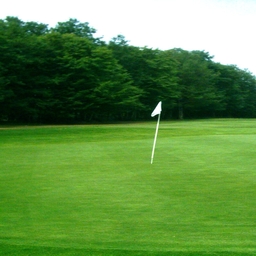} &
\includegraphics[width=.08\textwidth]{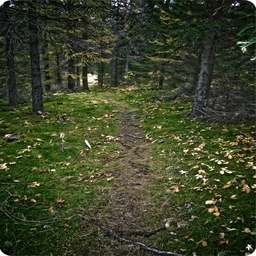} &
\includegraphics[width=.08\textwidth]{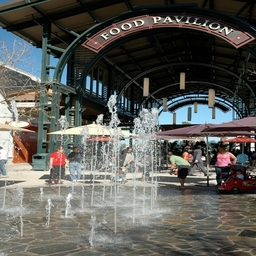} &
\includegraphics[width=.08\textwidth]{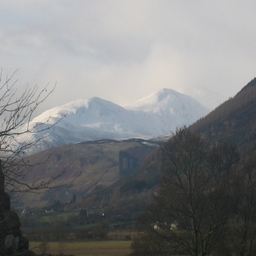}
\\
airport terminal &
apt.\ building/outdoor & 
ballroom &
bar &
beauty salon &
bowling alley &
dock &
excavation &
fairway &
forest path &
market outdoor &
mountain snowy
\\
\includegraphics[width=.08\textwidth]{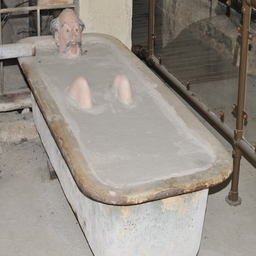} &
\includegraphics[width=.08\textwidth]{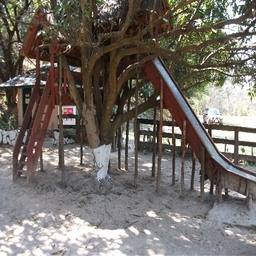} &
\includegraphics[width=.08\textwidth]{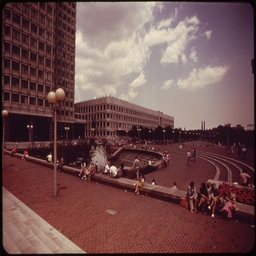} &
\includegraphics[width=.08\textwidth]{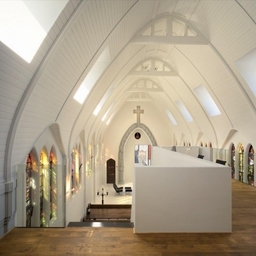} &
\includegraphics[width=.08\textwidth]{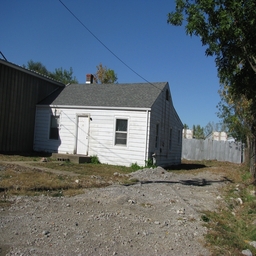} &
\includegraphics[width=.08\textwidth]{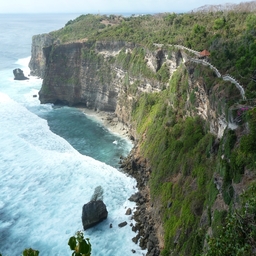} &
\includegraphics[width=.08\textwidth]{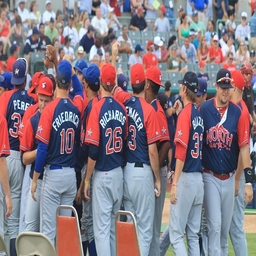} &
\includegraphics[width=.08\textwidth]{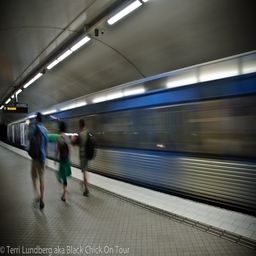} &
\includegraphics[width=.08\textwidth]{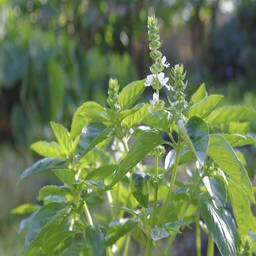} &
\includegraphics[width=.08\textwidth]{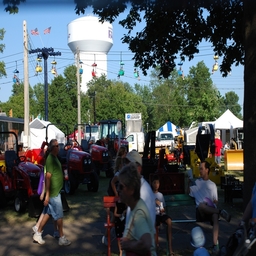} &
\includegraphics[width=.08\textwidth]{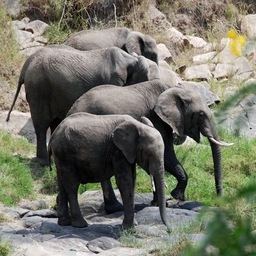} &
\includegraphics[width=.08\textwidth]{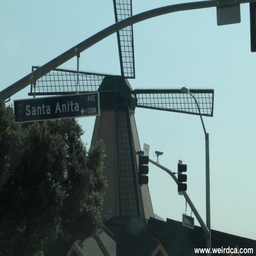}
\\
museum indoor &
picnic area &
plaza &
pulpit &
residential &  
sea cliff &
stadium/baseball &
subway platform &  
vegetable garden &
water tower &
watering hole &
windmill
\end{tabular}
\egroup
\vspace{1 mm}
\caption{Random images from 24 of the 205 classes in Places-205.}
\label{fig:places-samples}
\end{figure*}

\subsection{Adaptation}

\begin{figure*}[t]
\centering
\bgroup
\setlength{\tabcolsep}{16pt}
\begin{tabular}{ccc}
\includegraphics[width=.25\textwidth]{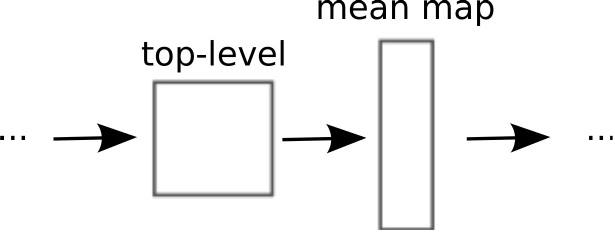}  &
\includegraphics[width=.25\textwidth]{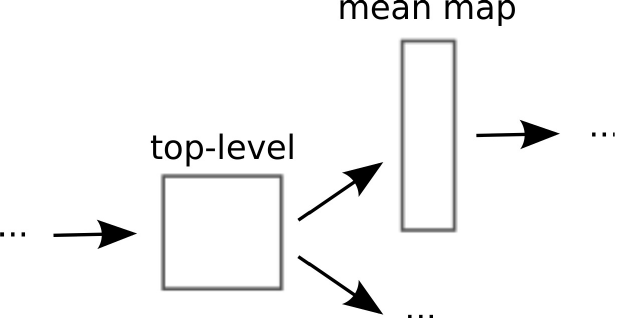} &
\includegraphics[width=.25\textwidth]{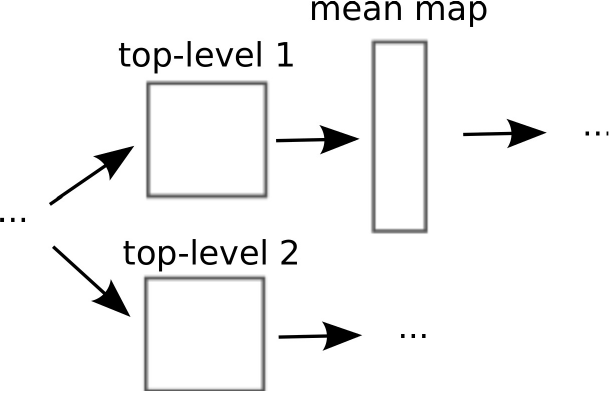} 
\\
Replacing &
Replicating &
Forking
\\
\end{tabular}
\egroup
\vspace{1 mm}
\caption{Illustration of each method for extending existing deep models. Top-level refers to the feature from the last convolutions layer. \textit{Replacing} here we append a mean map layer to the top-level convolutional features. \textit{Replicating} here we copy the top-level convolutional features and feed one copy to the mean map layer while passing the other copy through to classification layers. \textit{Forking} here we make two distinct top-level convolutional features, one that is fed to a mean map layer and the other is fed through to classification layers.  }
\label{fig:adapt}
\end{figure*}

In this section we illustrate the effectiveness of deep mean maps on several real-world scene classification datasets. We shall begin with a network trained on the ImageNet dataset \cite{russakovsky2014imagenet} and fine-tune it for each dataset.

\subsubsection{Datasets}

\paragraph{Flickr Style}
The first dataset we consider, known as Flickr Style,
was assembled by \cite{karayev:image-style}.
It contains 80,000 images in 20 style categories
defined by user groups
such as ``Bright,'' ``Noir,'' ``Long Exposure'', ``Geometric,'' and ``Romantic.''
We use the standard 80-20 train-test split,
and further divide the training images with an 80-20 train-validation split
for model selection.
Figure~\ref{fig:flickr-samples} shows an image from each class.

\paragraph{Wikipaintings}
This dataset, also assembled by \cite{karayev:image-style},
contains about 83,000 images
in 25 genres
including ``Impressionism,'' ``Baroque,'' ``Surrealism,'' and ``Cubism.''
Each category contains at least 1,000 images.
We divide the dataset randomly into
53,140 training,
13,285 validation,
and 16,612 test images.
Figure~\ref{fig:wikipaintings} shows an image from each class.

\paragraph{Places}
The Places-205 dataset \cite{zhou:places}
contains 2,448,873 training, 20,500 validation, and 41,000 test images
in 205 scene categories,
including
``runway,''
``ski resort,''
``trench,''
``kasbah,''
and ``dorm room.''
The test labels are not public; submissions must be scored on a server run by the authors of \cite{zhou:places}.
Figure~\ref{fig:places-samples} shows sample images.

\subsubsection{Networks}
We consider both a minor variation of the model of \cite{krizhevsky2012imagenet}
known as CaffeNet (\texttt{bvlc\_reference\_caffenet}),
and the GoogLeNet model of \cite{szegedy:googlenet} (\texttt{bvlc\_googlenet}). We fine-tune each network with SGD, with a consistent learning rate schedule for each architecture, and pick the model of our occasional snapshots with highest validation accuracy. Training time for each variant is comparable to the original network.

We extend these networks to use mean map layers (with 4,096 frequencies) in the following ways:
\begin{description}
  \item[Replacing] By replacing the fully connected layers with a mean map layer after the top-level features of the CNN, we look to classify solely on the mean map embedding of these top level features.
  \item[Replicating] One may send the top-level features both to a hidden layer as well as to a mean map layer; this allows one to classify with a combination of the vectorized top-level features and the mean map embedding.
  \item[Forking] Furthermore, rather than sending the same top-level features into the hidden and mean map layers above as in replication, one may learn two sets of distinct top-level-features to be fed to a hidden and mean map layer respectively.
\end{description}

We also consider the following variants on each model:
\begin{description}
  \item[Dropout] Add a dropout layer after the mean map embedding. This may encourage the network to avoid relying on any one frequency in the random embedding, thereby more faithfully replicating the kernel.
  \item[Hidden] Add a fully-connected hidden layer between the mean map embedding and the classifier.
  \item[Frequency Learning] Optimize the individual frequencies sampled, rather than just their scale.
\end{description}

Model selection was performed on the validation set; the three best MME models, along with the original model, were each scored on the test set.

\begin{table*}[p!]
\centering
\begin{tabular}{cccccc}
Architecture & DMM Type & Variants & Validation Top-1 & Test Top-1 & Test Top-3 \\
\hline\hline
CaffeNet & None & - & 37.69\% & 36.98\% & 62.17\% \\
CaffeNet & Replacing & Dropout, Hidden, Frequency Learning & \textbf{38.87\%} & \textbf{38.61\%} & \textbf{64.47\%} \\
CaffeNet & Replacing & Dropout, Hidden & 38.27\% & 37.91\% & 63.64\% \\
CaffeNet & Replicating & Dropout, Hidden & 38.27\% & 37.71\% & 63.31\% \\ \hline
GoogLeNet & None & - & 40.44\% & 39.93\% & 66.32\% \\
GoogLeNet & Replicating & Dropout, Frequency Learning & \textbf{40.74\%} & \textbf{40.20\%} & \textbf{66.66\%} \\
GoogLeNet & Replicating & Dropout & 40.66\% & 40.10\% & 66.26\% \\
GoogLeNet & Replicating & Dropout, Hidden, Frequency Learning & 40.48\% & 39.66\% & 66.32\% \\
\end{tabular}
\vspace{1 mm}
\caption{Adaptation results on the Flickr Style dataset. \textbf{Boldface} indicates best score for CaffeNet and GoogLeNet.}
\label{tab:flickr_style}
\end{table*}

\begin{table*}[p!]
\centering
\begin{tabular}{cccccc}
Architecture & DMM Type & Variants & Validation Top-1 & Test Top-1 & Test Top-3 \\
\hline\hline
CaffeNet & None & -             & 55.96\% & 54.94\% & 79.99\% \\
CaffeNet & Replacing & -     & \textbf{56.03\%} & \textbf{55.16\%} & \textbf{80.63\%} \\
CaffeNet & Forking & Dropout & 55.92\% & 55.03\% & 80.29\% \\
CaffeNet & Forking & -       & 55.76\% & 54.91\% & 80.45\% \\ \hline
GoogLeNet & None & -                                          & \textbf{58.94}\% & 57.49\% & 83.72\% \\
GoogLeNet & Replicating & Dropout                          & 58.18\% & 57.06\% & 83.64\% \\
GoogLeNet & Replicating & Dropout, Frequency Learning         & 57.96\% & \textbf{57.56\%} & \textbf{83.84\%} \\
GoogLeNet & Replicating & Dropout, Hidden, Frequency Learning & 57.70\% & 56.36\% & 83.25\% \\
\end{tabular}
\vspace{1 mm}
\caption{Adaptation results on the Wikipaintings dataset. \textbf{Boldface} indicates best score for CaffeNet and GoogLeNet.}
\label{tab:wikipaintings}
\end{table*}


\begin{table*}[p!]
\centering
\begin{tabular}{cccccc}
Architecture & DMM Type & Variants & Validation Top-1 & Test Top-1 & Test Top-5  \\
\hline\hline
CaffeNet & None & -                 & 45.92\% & 45.69\% & 77.45\% \\
CaffeNet & Forking & -           & \textbf{46.86\%} & \textbf{46.62\%} & \textbf{77.99\%} \\
CaffeNet & Replicating & -       & 46.40\% & 46.31\% & 77.97\% \\
CaffeNet & Forking & Dropout     & 46.19\% & 46.27\% & 77.95\% \\ \hline
GoogLeNet & None & -                                      & 47.50\% & 47.50\% & 79.83\% \\
GoogLeNet & Forking & Dropout, Hidden                  & \textbf{49.17\%} & 48.99\% & \textbf{80.45\%} \\
GoogLeNet & Forking & Dropout, Hidden, Frequency Learning & 48.80\% & \textbf{49.00\%} & 80.42\% \\
GoogLeNet & Forking & Dropout,         Frequency Learning & 48.32\% & 48.56\% & 80.21\%
\end{tabular}
\vspace{1 mm}
\caption{Adaptation results on the Places dataset. \textbf{Boldface} indicates best score for CaffeNet and GoogLeNet.}
\label{tab:places}
\end{table*}

\subsubsection{Results}

Tables~\ref{tab:flickr_style}-\ref{tab:places} present adaptation results for each of the three datasets for both CaffeNet and GoogleNet with the three best DMM extensions (DMM type) on the validation set and the ``base'' model (DMM type ``None''). 

The best model variant picked according to validation score nearly always outperform the equivalent ``base'' model on the test set on Flickr Style and Wikipaintings. Hence we see that DMMs, through the use on distributions of top-level features, are able top improve existing deep-architectures. Furthermore, while the flexible nature of the mean map layer allows it to extend existing networks in many different variations, DMMs seem to generalize well empirically. Thus, a simple validation set approach is effective for choosing how to extend a network's architecture with DMMs.

\section{Conclusion} \label{sec:conclusion}

\subsection{Related Work}
As previously mentioned, deep architectures for learning high-level features has been explored in a myriad of work \cite{hinton2006reducing,lecun2012learning,krizhevsky2012imagenet,lee2009convolutional,le2013building}. While such high-level features may be extracted in a variety of different deep architectures, we focus on deep convolution neural networks (CNNs) \cite{lecun1995convolutional,lecun1998gradient,lecun2004learning,lecun2012learning}. Modern CNNs have been able to employ more layers \cite{krizhevsky2012imagenet, szegedy2014going} without unsupervised pretraining thanks to their reduced parameterization (as compared to general neural networks), recent advents in training methods and nonlinear units \cite{hinton2012improving,nair2010rectified}, and large datasets \cite{russakovsky2014imagenet}. 

Also, a great deal work has been devoted to constructing distribution-based features for vision tasks. These distribution-based approaches are often histograms such as BoW representations  \cite{fei2005bayesian,leung2001representing}, HoG features \cite{dalal2005histograms}, and dense-SIFT \cite{lowe1999object}. Moreover, there have been efforts to extended distribution-based methods to nonparametric continuous domains \cite{poczos2012nonparametric}. 

There has been work that explores the use of distributions of high-level deep architecture features with ad-hoc features from distributions of fixed, pre-trained high-level features \cite{wu2015visual}. In contrast, this paper learns high-level features and uses their distributions in a scalable, nonparametric fashion with mean maps.

Mean map embeddings \cite{smola2007hilbert,gretton2006kernel,song2008learning} have served as a practical and flexible method to learn using datasets of sample sets drawn from distributions \cite{muandet2012learning,szabo2014learning}. Furthermore, the use of random features \cite{rahimi2007random,rahimi2009weighted} has allowed for scalable learning over distributions \cite{oliva2013fast,jitkrittum2015kernel,lopez2015towards,flaxman2015supported}. Mean map embeddings have also been used extensively for the comparison and two-sample testing of distributions \cite{gretton2006kernel,gretton2012kernel,fukumizu2007kernel}.
Recent research has also explored the use of mean maps in deep architectures for building fast sampling mechanisms, by conducting two-sample tests between generated sample points and original data \cite{li2015generative,dziugaite2015training}; this work differs from ours in using mean maps in the loss function, rather than in the architecture itself.

\subsection{Remarks}
This paper presents DMMs, a novel framework to jointly use and learn distributions of top-level features in deep architectures. We showed that DMMs, though a mean map layer, may use the mean embedding of top-level features to non-parametrically represent the distributions of the features. 

The use of mean embeddings with random features allow the mean map layer to non-parametrically represent distributions of top-level features whilst still scaling to large image datasets. Also, inner-products on the mean embeddings through the network are interpretable as RKHS inner products on the distributional embeddings, allowing one to build a strong theoretical foundation \cite{szabo2014learning}. Furthermore, we showed that the mean map layer may be implemented using typical CNN operations, making both forward and backward propagation simple to do with DMMs. 

Moreover, we illustrated the aptitude of the mean map layer at learning distributions of visual features for discrimination in a synthetic data experiment. We saw that even with very few instances the mean map layer allowed a network to quickly learn to distinguish visual distributional patterns in a sample efficient manner. Lastly, we showed that DMMs may be used to extend several existing state-of-the-art deep-architectures and improve their performance on various challenging real-world datasets. Indeed, the mean map layer prove flexible and capable of extending networks an a variety of ways; due to a propensity to generalize well, it is simple to choose an extension method with a straightforward validation set approach. Thus, it is clear that the DMM framework can successfully build on the myriad of state-of-the-art deep architectures available.

\clearpage

{\small
\bibliographystyle{ieee}
\bibliography{bib}

\begin{thebibliography}{10}\itemsep=-1pt

\bibitem{textbrodatz}
Colored brodatz texture database.
\newblock
  \url{http://multibandtexture.recherche.usherbrooke.ca/colored\%20_brodatz.html}.

\bibitem{dalal2005histograms}
N.~Dalal and B.~Triggs.
\newblock Histograms of oriented gradients for human detection.
\newblock In {\em Computer Vision and Pattern Recognition, 2005. CVPR 2005.
  IEEE Computer Society Conference on}, volume~1, pages 886--893. IEEE, 2005.

\bibitem{dziugaite2015training}
G.~K. Dziugaite, D.~M. Roy, and Z.~Ghahramani.
\newblock Training generative neural networks via maximum mean discrepancy
  optimization.
\newblock {\em arXiv:1505.03906}, 2015.

\bibitem{fei2005bayesian}
L.~Fei-Fei and P.~Perona.
\newblock A {B}ayesian hierarchical model for learning natural scene
  categories.
\newblock In {\em Computer Vision and Pattern Recognition, 2005. CVPR 2005.
  IEEE Computer Society Conference on}, volume~2, pages 524--531. IEEE, 2005.

\bibitem{flaxman2015supported}
S.~R. Flaxman, Y.-X. Wang, and A.~J. Smola.
\newblock Who supported {O}bama in 2012?: Ecological inference through
  distribution regression.
\newblock In {\em Proceedings of the 21th ACM SIGKDD International Conference
  on Knowledge Discovery and Data Mining}, pages 289--298. ACM, 2015.

\bibitem{fukumizu2007kernel}
K.~Fukumizu, A.~Gretton, X.~Sun, and B.~Sch{\"o}lkopf.
\newblock Kernel measures of conditional dependence.
\newblock In {\em NIPS}, volume~20, pages 489--496, 2007.

\bibitem{gretton2006kernel}
A.~Gretton, K.~M. Borgwardt, M.~Rasch, B.~Sch{\"o}lkopf, and A.~J. Smola.
\newblock A kernel method for the two-sample-problem.
\newblock In {\em Advances in neural information processing systems}, pages
  513--520, 2006.

\bibitem{gretton2012kernel}
A.~Gretton, K.~M. Borgwardt, M.~J. Rasch, B.~Sch{\"o}lkopf, and A.~Smola.
\newblock A kernel two-sample test.
\newblock {\em The Journal of Machine Learning Research}, 13(1):723--773, 2012.

\bibitem{hinton2006reducing}
G.~E. Hinton and R.~R. Salakhutdinov.
\newblock Reducing the dimensionality of data with neural networks.
\newblock {\em Science}, 313(5786):504--507, 2006.

\bibitem{hinton2012improving}
G.~E. Hinton, N.~Srivastava, A.~Krizhevsky, I.~Sutskever, and R.~R.
  Salakhutdinov.
\newblock Improving neural networks by preventing co-adaptation of feature
  detectors.
\newblock {\em arXiv preprint arXiv:1207.0580}, 2012.

\bibitem{caffe}
Y.~Jia, E.~Shelhamer, J.~Donahue, S.~Karayev, J.~Long, R.~Girshick,
  S.~Guadarrama, and T.~Darrell.
\newblock Caffe: Convolutional architecture for fast feature embedding.
\newblock {\em arXiv:1408.5093}, 2014.

\bibitem{jitkrittum2015kernel}
W.~Jitkrittum, A.~Gretton, N.~Heess, S.~Eslami, B.~Lakshminarayanan,
  D.~Sejdinovic, and Z.~Szab{\'o}.
\newblock Kernel-based just-in-time learning for passing expectation
  propagation messages.
\newblock {\em arXiv:1503.02551}, 2015.

\bibitem{karayev:image-style}
S.~Karayev, M.~Trentacoste, H.~Han, A.~Agarwala, T.~Darrell, A.~Hertzmann, and
  H.~Winnemoeller.
\newblock Recognizing image style.
\newblock {\em arXiv:1311.3715}, 2013.

\bibitem{krizhevsky2012imagenet}
A.~Krizhevsky, I.~Sutskever, and G.~E. Hinton.
\newblock Imagenet classification with deep convolutional neural networks.
\newblock {\em Advances in neural information processing systems}, pages
  1097--1105, 2012.

\bibitem{le2013building}
Q.~V. Le.
\newblock Building high-level features using large scale unsupervised learning.

\bibitem{lecun2012learning}
Y.~LeCun.
\newblock Learning invariant feature hierarchies.
\newblock In {\em Computer vision--ECCV 2012. Workshops and demonstrations},
  pages 496--505. Springer, 2012.

\bibitem{lecun1995convolutional}
Y.~LeCun and Y.~Bengio.
\newblock Convolutional networks for images, speech, and time series.
\newblock {\em The handbook of brain theory and neural networks}, 3361(10),
  1995.

\bibitem{lecun1998gradient}
Y.~LeCun, L.~Bottou, Y.~Bengio, and P.~Haffner.
\newblock Gradient-based learning applied to document recognition.
\newblock {\em Proceedings of the IEEE}, 86(11):2278--2324, 1998.

\bibitem{lecun2004learning}
Y.~LeCun, F.~J. Huang, and L.~Bottou.
\newblock Learning methods for generic object recognition with invariance to
  pose and lighting.
\newblock In {\em Computer Vision and Pattern Recognition, 2004. CVPR 2004.
  Proceedings of the 2004 IEEE Computer Society Conference on}, volume~2, pages
  II--97. IEEE, 2004.

\bibitem{lecun2013prez}
Y.~LeCun and M.~Ranzato.
\newblock Deep learning tutorial.
\newblock International Conference on Machine Learning, 2013.

\bibitem{lee2009convolutional}
H.~Lee, R.~Grosse, R.~Ranganath, and A.~Y. Ng.
\newblock Convolutional deep belief networks for scalable unsupervised learning
  of hierarchical representations.
\newblock In {\em Proceedings of the 26th Annual International Conference on
  Machine Learning}, pages 609--616. ACM, 2009.

\bibitem{leung2001representing}
T.~Leung and J.~Malik.
\newblock Representing and recognizing the visual appearance of materials using
  three-dimensional textons.
\newblock {\em International journal of computer vision}, 43(1):29--44, 2001.

\bibitem{li2015generative}
Y.~Li, K.~Swersky, and R.~Zemel.
\newblock Generative moment matching networks.
\newblock {\em arXiv:1502.02761}, 2015.

\bibitem{lopez2015towards}
D.~Lopez-Paz, K.~Muandet, B.~Sch{\"o}lkopf, and I.~Tolstikhin.
\newblock Towards a learning theory of causation.
\newblock {\em arXiv:1502.02398}, 2015.

\bibitem{lowe1999object}
D.~G. Lowe.
\newblock Object recognition from local scale-invariant features.
\newblock In {\em Computer vision, 1999. The proceedings of the seventh IEEE
  international conference on}, volume~2, pages 1150--1157. Ieee, 1999.

\bibitem{lin2014nin}
S.~Y. M.~Lin, Q.~Chen.
\newblock Network in network.

\bibitem{muandet2012learning}
K.~Muandet, K.~Fukumizu, F.~Dinuzzo, and B.~Sch{\"o}lkopf.
\newblock Learning from distributions via support measure machines.
\newblock {\em Advances in neural information processing systems}, pages
  10--18, 2012.

\bibitem{nair2010rectified}
V.~Nair and G.~E. Hinton.
\newblock Rectified linear units improve restricted boltzmann machines.
\newblock In {\em Proceedings of the 27th International Conference on Machine
  Learning (ICML-10)}, pages 807--814, 2010.

\bibitem{oliva2013fast}
J.~B. Oliva, W.~Neiswanger, B.~P{\'o}czos, J.~Schneider, and E.~Xing.
\newblock Fast distribution to real regression.
\newblock {\em arXiv preprint arXiv:1311.2236}, 2013.

\bibitem{poczos2012nonparametric}
B.~P{\'o}czos, L.~Xiong, D.~J. Sutherland, and J.~Schneider.
\newblock Nonparametric kernel estimators for image classification.
\newblock In {\em Computer Vision and Pattern Recognition (CVPR), 2012 IEEE
  Conference on}, pages 2989--2996. IEEE, 2012.

\bibitem{rahimi2007random}
A.~Rahimi and B.~Recht.
\newblock Random features for large-scale kernel machines.
\newblock {\em Advances in neural information processing systems}, pages
  1177--1184, 2007.

\bibitem{rahimi2009weighted}
A.~Rahimi and B.~Recht.
\newblock Weighted sums of random kitchen sinks: Replacing minimization with
  randomization in learning.
\newblock In {\em Advances in neural information processing systems}, pages
  1313--1320, 2009.

\bibitem{russakovsky2014imagenet}
O.~Russakovsky, J.~Deng, H.~Su, J.~Krause, S.~Satheesh, S.~Ma, Z.~Huang,
  A.~Karpathy, A.~Khosla, M.~Bernstein, et~al.
\newblock Imagenet large scale visual recognition challenge.
\newblock {\em International Journal of Computer Vision}, pages 1--42, 2014.

\bibitem{smola2007hilbert}
A.~Smola, A.~Gretton, L.~Song, and B.~Sch{\"o}lkopf.
\newblock A hilbert space embedding for distributions.
\newblock In {\em Algorithmic Learning Theory}, pages 13--31. Springer, 2007.

\bibitem{song2008learning}
L.~Song, K.~Fukumizu, E.~Gordon, A.~Gretton, S.~Hong, B.~Sch{\"o}lkopf,
  E.~Smola, V.~Vishwanasan, and L.~Williams.
\newblock Learning via hilbert space embedding of distributions.
\newblock 2008.

\bibitem{szabo2014learning}
Z.~Szab{\'o}, A.~Gretton, B.~P{\'o}czos, and B.~Sriperumbudur.
\newblock Learning theory for distribution regression.
\newblock {\em arXiv preprint arXiv:1411.2066}, 2014.

\bibitem{szegedy2014going}
C.~Szegedy, W.~Liu, Y.~Jia, P.~Sermanet, S.~Reed, D.~Anguelov, D.~Erhan,
  V.~Vanhoucke, and A.~Rabinovich.
\newblock Going deeper with convolutions.
\newblock {\em arXiv preprint arXiv:1409.4842}.

\bibitem{szegedy:googlenet}
C.~Szegedy, W.~Liu, Y.~Jia, P.~Sermanet, S.~Reed, D.~Anguelov, D.~Erhan,
  V.~Vanhoucke, and A.~Rabinovich.
\newblock Going deeper with convolutions.
\newblock {\em arXiv:1409.4842}, 2014.

\bibitem{wu2015visual}
J.~Wu, B.-B. Gao, and G.~Liu.
\newblock Visual recognition using directional distribution distance.
\newblock {\em arXiv preprint arXiv:1504.04792}, 2015.

\bibitem{zhou:places}
B.~Zhou, A.~Lapedriza, J.~Xiao, A.~Torralba, and A.~Oliva.
\newblock Learning deep features for scene recognition using places database.
\newblock In {\em Advances in Neural Information Processing Systems}, 2014.

\end{thebibliography}
}

\end{document}